\documentclass[final]{cvpr}

\usepackage{times}
\usepackage{epsfig}
\usepackage{graphicx}
\usepackage{subfig}
\usepackage{amsmath}
\usepackage{amssymb}
\usepackage[english]{babel}
\usepackage{cuted}
\usepackage{capt-of}
\usepackage{placeins}
\usepackage[flushleft]{threeparttable}
\usepackage{mathtools}
\usepackage{textcomp}
\usepackage{gensymb}
\usepackage{subfig}
\usepackage{mwe}
\usepackage[numbers]{natbib}
\usepackage{bm}

\newlength{\tempheight}
\newlength{\tempwidth}

\newcommand{\rowname}[1]
{{\makebox[\tempheight][c]{\textbf{#1}}}}

\newcommand{\columnname}[1]
{\makebox[\tempwidth][c]{\textbf{#1}}}

\newcommand{\bracedincludegraphics}[2][]{%
  \sbox0{$\vcenter{\hbox{\includegraphics[#1]{#2}}}$}%
  \left\lbrace
    \vphantom{\copy0}
  \right.\kern-\nulldelimiterspace
  \underbrace{\box0}}
  
\newcommand{\bb}[1]{\bm{\mathrm{#1}}}

\usepackage[pagebackref=true,breaklinks=true,colorlinks,bookmarks=false]{hyperref}

\usepackage[capitalize]{cleveref}
\crefname{section}{Sec.}{Secs.}
\Crefname{section}{Section}{Sections}
\Crefname{table}{Table}{Tables}
\crefname{table}{Tab.}{Tabs.}

\def\cvprPaperID{6469} 
\def\confYear{CVPR 2022}

\begin{document}

\title{Delta-GAN-Encoder: Encoding Semantic Changes for Explicit Image Editing, using Few Synthetic Samples.}

\author{Nir Diamant \qquad Nitsan Sandor \qquad Alex M. Bronstein \\\\Technion - Israel Institute of Technology \\ \tt\small {\{nirdiamant21,nits210\}@gmail.com}}

\maketitle
\begin{strip}\centering
    \centering
    \includegraphics[width =0.83\textwidth ]{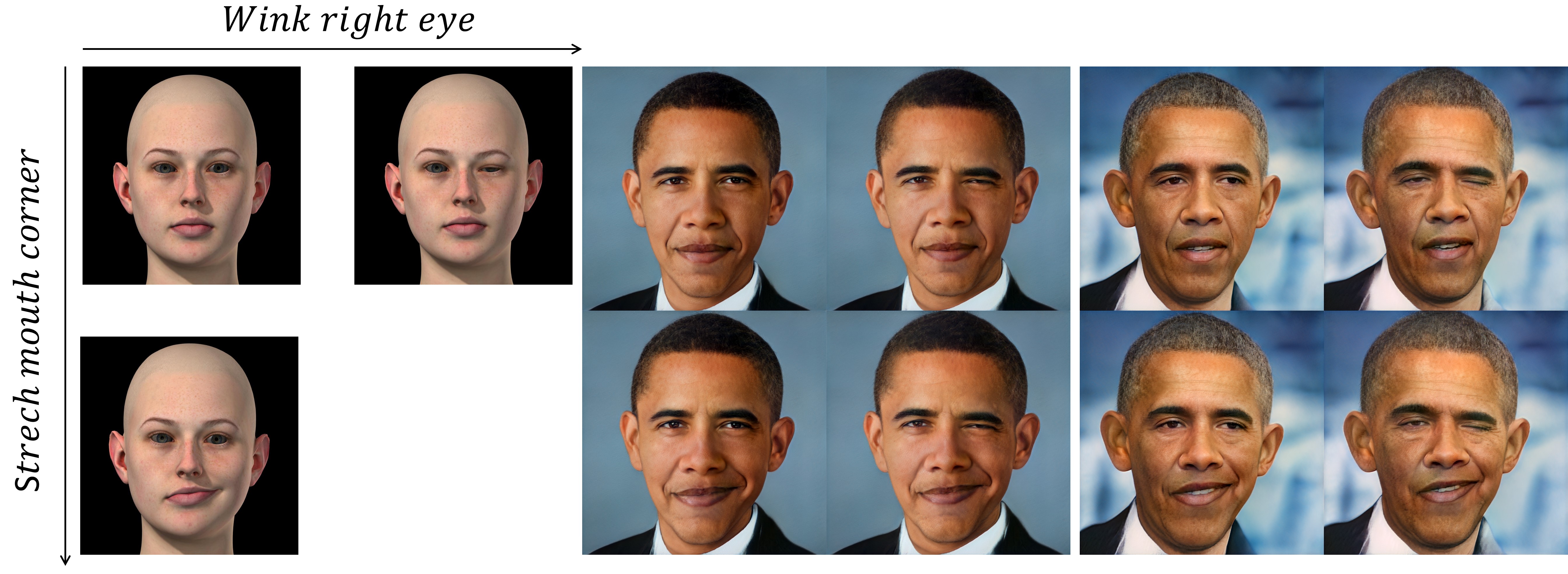}
    \captionof{figure}{An example of applying disentangled explicit changes of winking and stretching of the mouth corner, learning from a synthetic face for inferring on real ones. Each axis demonstrates movement in a learned nonlinear path of a GAN's latent space. The bottom right image in each quadrant is a composite of both changes, one superimposed on top of the another.
    \label{fig:title-fig}}

    \end{strip}
\begin{abstract}
  Understating and controlling generative models' latent space is a complex task.
  In this paper, we propose a novel method for learning to control any desired attribute in a pre-trained GAN's latent space, for the purpose of editing synthesized and real-world data samples accordingly.
  We perform Sim2Real learning, relying on minimal samples to achieve an unlimited amount of continuous precise edits.
  We present an Autoencoder-based model that learns to encode the semantics of changes between images as a basis for editing new samples later on, achieving precise desired results - example shown in Fig. \ref{fig:title-fig}.
  While previous editing methods rely on a known structure of latent spaces (e.g., linearity of some semantics in StyleGAN), our method inherently does not require any structural constraints.  
  We demonstrate our method in the domain of facial imagery: editing different expressions, poses, and lighting attributes, achieving state-of-the-art results.

\end{abstract}

\section{Introduction}
    
    Over the last few years, the concept of Generative Adversarial Networks (GANs) has been developing rapidly and improved the ability to synthesize more photo-realistic data samples, specifically images.
    Even though recent GANs' architectures are designed to have semantic latent space nature (e.g., StyleGAN variants \cite{karras2019stylebased, karras2020analyzing, karras2020training}), explicit control over the latent space has not yet been achieved.
    StyleGAN's mapping network $f(\bb{z})$ that maps the input latent vector $\bb{z}$ to an intermediate latent vector $\bb{W}$ contributes to disentangling the $W$-space, since it does not have to match the probability density of the training data as $\bb{z}$. Additionally, a disentangled representation will ease the generation of realistic samples \cite{karras2019stylebased,shen2020interpreting, jahanian2020steerability, tewari2020stylerig}.
   However, the different $W$-space semantics are encoded in an imperfectly linearly-separated manner, introducing difficulties in generating a specific desired output.
   
    CGANs \cite{mirza2014conditional} can overcome the problem of attribute entanglement, since it is trained to separate outputs. Despite this, CGANs are limited to features chosen pre-training. Another major drawback to achieving feature disentangled results is the imprecise control over the GAN output: the lack of an accurate metric for distinguishing between any two possible attributes prohibits the generation of every specific pre-defined image.     
    The existence of a real-world derived dataset containing the same samples differing only by specific attributes may contribute to the definition and inference of exact attributes, though this is nearly impossible to achieve. 
    
    Studies have been conducted on how to project real-world images into their matching latent vectors to learn and control the behavior of latent spaces.
    For StyleGAN, it is common to project an image into $\bb{W^+}$ vectors  \cite{abdal2019image2stylegan, abdal2020image2stylegan}, which are concatenation of 18 different $\bb{W}$ vectors.
    Nevertheless, projections into a GAN's latent space still cannot mimic the image generation's inverse function, and exhibit a known issue of the trade-off between qualitative image reconstruction and the latent vector's semantics. 
  
  Generative Normalizing Flows \cite{Kobyzev_2020, kingma2018glow, hoogeboom2019emerging, izmailov2019semisupervised} can perfectly project an image by calculating the inverse of the generative function thanks to their property of reversibility, since they are composed of a series of bijective functions. A significant disadvantage is that the latent vector size must match the image dimensions, producing a high-dimensional latent space whose exploration time is longer by orders of magnitude.
  
  Recent work by Schwartz \etal \cite{schwartz2018deltaencoder} proposes a method of encoding changes between images in the feature space using an Autoencoder-based model. The encoded changes are then used for augmenting sparse classes and improving classification accuracy.
  
  Deriving from Schwartz \etal \cite{schwartz2018deltaencoder}, we propose the idea of encoding the changes ($\bb{\Delta}$-s) and put forth a method that enhances this concept for generating accurate edited qualitative data samples using a pre-trained GAN. To achieve this goal, we present a new method requiring three crucial elements: a qualitative projection method (i.e., reconstruction and editability), multiple samples differing by only a single attribute, and a model that can encode the changes and apply them to edit latent vectors representing images.

Our contribution is in providing a method that:
\begin{enumerate}
    \item Allows the editing of images with unlimited specificity unencumbered by attribute entanglement, learning to isolate the different semantics.
    \item Learns the behavior of the semantics without the requirement of any structural constraints, allowing it to find the precise nonlinear path of the edit in the latent space.
\end{enumerate}
Our code and data will be available to download from www.xxxxxxx.com.

\section{Related Work}
    Different projects have been completed towards achieving the disentanglement of features in generative models \cite{liu2020oogan, eom2019learning, 8099624, ren2021generative}.
    Shoshan \etal \cite{shoshan2021gancontrol} proposed using pre-trained task-oriented models to reward and penalize every two samples that share or differ by specific criteria, thus clustering the semantics within the latent space structure. Shen \etal \cite{8578190} added a face classifier as a third player to compete against the Generator in order to sharpen its ability to preserve the identity of the generated image. 
    
    Learning from synthetic data and combining 3D models with GANs were used for multi-domain applications (e.g., expression transfer, pose changing, 3D face out of 2D face constructing, face de-occlusion, face frontalization, face decoupling, etc.) \cite{geng20193d, mokhayeri2019crossdomain, Gecer_2019, yuan2019face, yin2017largepose, abrevaya2019decoupled}.
   
    Some works have leveraged the relatively linear separable structure of StyleGAN's latent space and explored the semantics of linear differences between latent vectors
    \cite{abdal2019image2stylegan, abdal2020image2stylegan}. Other studies were conducted to find meaningful linear paths in the latent space \cite{shen2020interpreting, wu2020stylespace, harkonen2020ganspace}.
    Zhu \etal \cite{zhu2021improved} studied where good latent codes are located in the latent space.
    
    Another related task is image projection into the latent space for editing real image samples using generative models. Some methods use iterative optimization algorithms \cite{karras2020analyzing, anirudh2020mimicgan}, while other methods use different kinds of Encoder-based architectures to project an image into its matching latent vector. The latter tend to reconstruct less accurately yet significantly more quickly, embedding the vector much more semantically. \cite{zhu2020indomain, richardson2021encoding, wei2021simple}.
 
    Tov \etal \cite{tov2021designing} designed an encoder that embeds images into semantically better latent vectors with a slight reconstruction quality trade-off. Alaluf \etal \cite{alaluf2021restyle} enhanced both encoders from \cite{richardson2021encoding, tov2021designing} by using their models iteratively and learning the residual with respect to the current estimate of the latent vector at each step until convergence.
    Alaluf \etal \cite{alaluf2021matter} studied the latent space semantics of aging transformation, supervised by an age prediction model, and showed that the actual latent path is nonlinear, surpassing the results of a linear path. Abdal \etal \cite{abdal2020styleflow} formulated conditional exploration in the latent space as an instance of conditional continuous normalizing flows to enhance attribute-conditioned sampling and attribute-controlled editing. They showed that the nonlinear paths surpass the quality of the results that correspond to the linear path.

\section{Method}

\subsection{General Idea}
\begin{figure*}[]
    \centering
    \includegraphics[width =0.83\textwidth ]{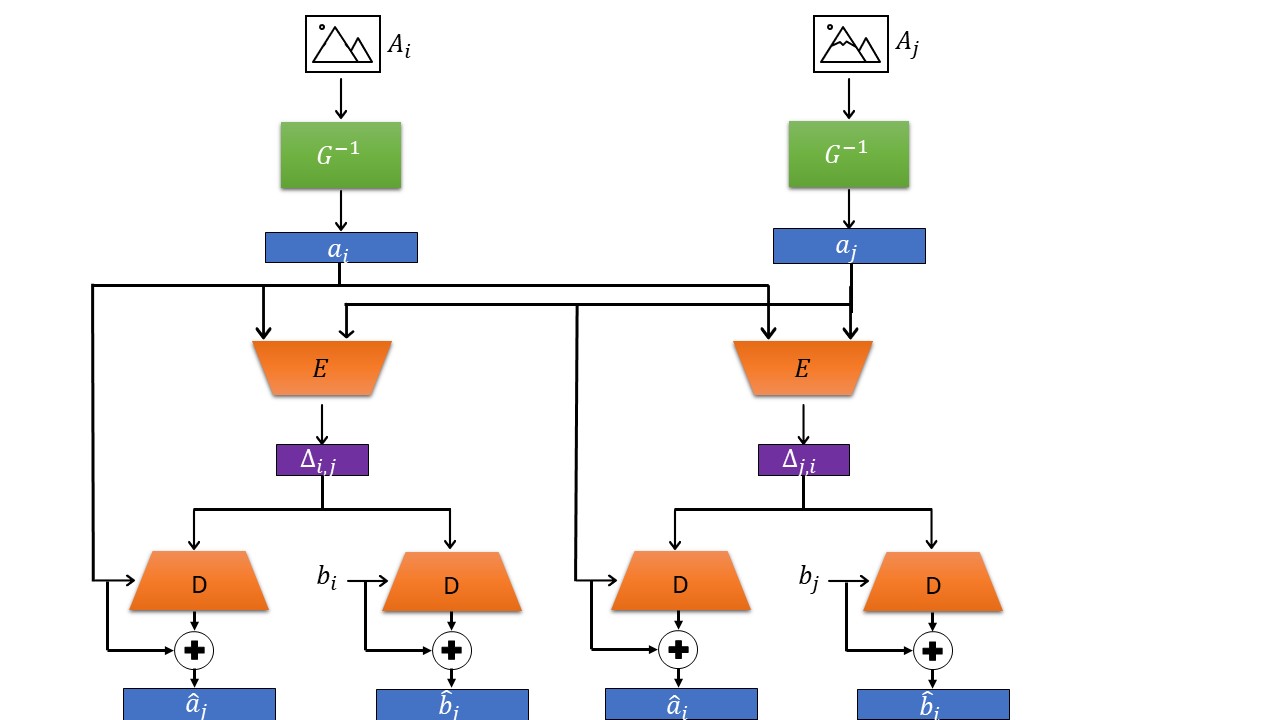}
    \caption{Our model's structure: Two images $\bb{A}_i, \bb{A}_j$ are projected into the GAN's latent vectors $\bb{a}_i$, $\bb{a}_j \in W^+$ with a semantic projection method $G^{-1}$ . Both latent vectors are then fed into the encoder $E$ to calculate a small vector, $\bb{\Delta}_{i,j}$, representing the directed semantic change between $\bb{a}_j$ and $\bb{a}_i$. $\bb{\Delta}_{i,j}$ is being fed into the decoder $D$ once with $\bb{a}_i$ and then with $\bb{b}_i$ which outputs the residual needed to get $\bb{a}_j$ and $\bb{b}_j$ respectively. The whole pipeline is symmetric for $\bb{\Delta}_{j,i}$.}
    \label{fig:delta-gan-encoder-arch}
\end{figure*}

 Our method finds the latent vector $\bb{a}_j \in W^+$ in a pre-trained GAN that matches the desired transformed image $\bb{A}_j$, given an input image $\bb{A}_i$. It does it by first learning $\bb{\Delta}_{i,j}$, a low dimensional vector that represents a semantic change defined by two images, and then combines it to the input vector $\bb{a}_i \in W^+$ to get the desired result. Both parts are learned together during the training phase of our model. Full model architecture is illustrated in Fig. \ref{fig:delta-gan-encoder-arch}.
 
  We assume that using projection methods that embed images semantically into $\bb{W^+}$ \cite{abdal2019image2stylegan, abdal2020image2stylegan} vectors will embed less-realistic images semantically similar to embedding real ones (e.g., the manipulation over the latent vector for editing the eyes of a natural person is similar to editing the eyes of a synthetic face).  This vital observation allows us to learn the task on synthetic data and infer the results over real-world samples.
 
\begin{figure}
     \centering
     \includegraphics[width=0.53\columnwidth ]{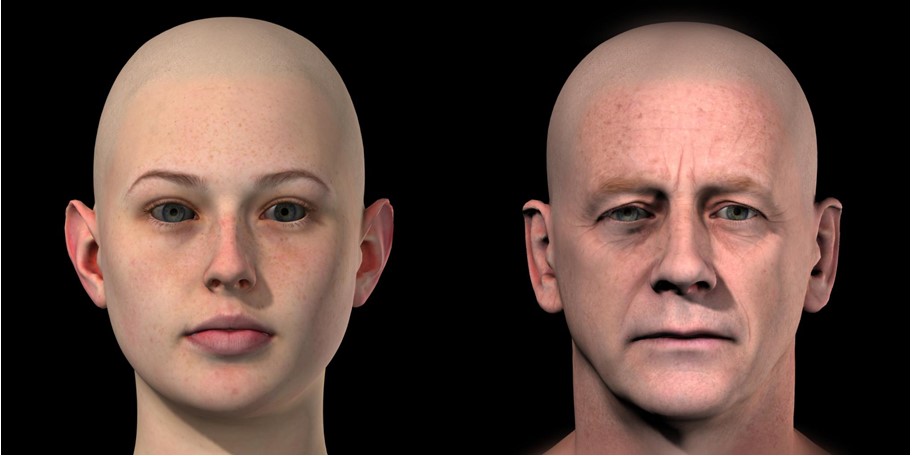}
     \caption{Female and Male synthetic 3D models.}
     \label{fig:synthetic_female_male}
 \end{figure}

 To supervise the process of learning the $\bb{\Delta}$-s, we create a small image dataset generated from synthetic 3D models of one female and one male - an example is shown in Fig. \ref{fig:synthetic_female_male}.
 The dataset consists of two classes, $A$ and $B$, for each we have n-sized sequence of samples, $\bb{A}_1, \bb{A}_2,..., \bb{A}_n$ and $\bb{B}_1, \bb{B}_2,...,\bb{B}_n$. In each series, one of its attributes changes along the sequence, while all the other attributes remain untouched. We then project all the $\bb{A}_i$ and $\bb{B}_i$ samples into their matching latent vectors $\bb{a}_i$, $\bb{b}_i$ according to the related pre-trained GAN. 
 
 \subsection{Model Architecture and Loss Function}
 Our model is based on an Autoencoder architecture, where
 the Encoder $E$ receives two latent vectors, $\bb{a}_i$, $\bb{a}_j$ $\in$ $W^+$, that match the images $\bb{A}_i$, $\bb{A}_j$. Both images belong to the same class (denoted alphabetically) and differ by some attribute (denoted by subscript). The model encodes them to $\bb{\Delta}_{i,j}$, a lower-dimensional vector that represents the semantic distinction between the inputs.
 \begin{equation}
     \bb{\Delta}_{i,j} = E(\bb{a}_i, \bb{a}_j).
 \end{equation}
 We denote the space, consisting of all such $\bb{\Delta}$-s as $\Delta$-space. We enforce linear space properties on the $\Delta$-space as described in subsection \ref{Enforcing linear space properties}
 
The Decoder $D$ receives a latent vector $\bb{b}_i$, that matches the image $\bb{B}_i$, and $\bb{\Delta}_{i,j}$. It produces an intermediate term $\hat{\bb{b}_j}_{residual}$ which is added to $\bb{b}_i$ to encourage the model to learn the residual of $\bb{b}_j$ and $\bb{b}_i$.
$D$ produces the residual $\hat{\bb{b}_j}_{residual}$ instead of $\hat{\bb{b}_j}$, since it is an easier task than learning the explicit output directly. 
\begin{equation}
    \hat{\bb{b}_j} = \underbrace{D(\bb{b}_i, \bb{\Delta}_{i,j})}_\text{$\hat{\bb{b}_j}_{residual}$} + \bb{b}_i.
\end{equation}
The process is done in a supervised manner using two losses: 
(1) Identity loss - the $\bb{\Delta}$ is defined and applied to the same class. (2) Transfer loss - the $\bb{\Delta}$ is defined by one class and applied to another.

\begin{equation}
l_{residual} = \underbrace{\lambda_{1}\cdot\lVert \bb{a}_j - \hat{\bb{a}_j}\rVert_2^2}_\text{loss over a seen class} + \underbrace{\lambda_{2}\cdot\lVert \bb{b}_j - \hat{\bb{b}_j}\rVert_2^2}_\text{loss over an unseen class}.
\end{equation}

The above explanation holds where the classes of $A, B$ and the indices $i, j$ change roles during training. This mechanism encourages the model to be class symmetric and emphasizes that the order of inputs to $E$ is important: opposite order of inputs, $(j, i)$ instead of $(i,j)$, yields an exact opposite $\bb{\Delta}_{j, i}$, a linear property that we force over the $\Delta$-space. \\

\subsection{Real Time Data Augmentation}
\begin{figure}[]
    \centering
    \includegraphics[width =\columnwidth ]{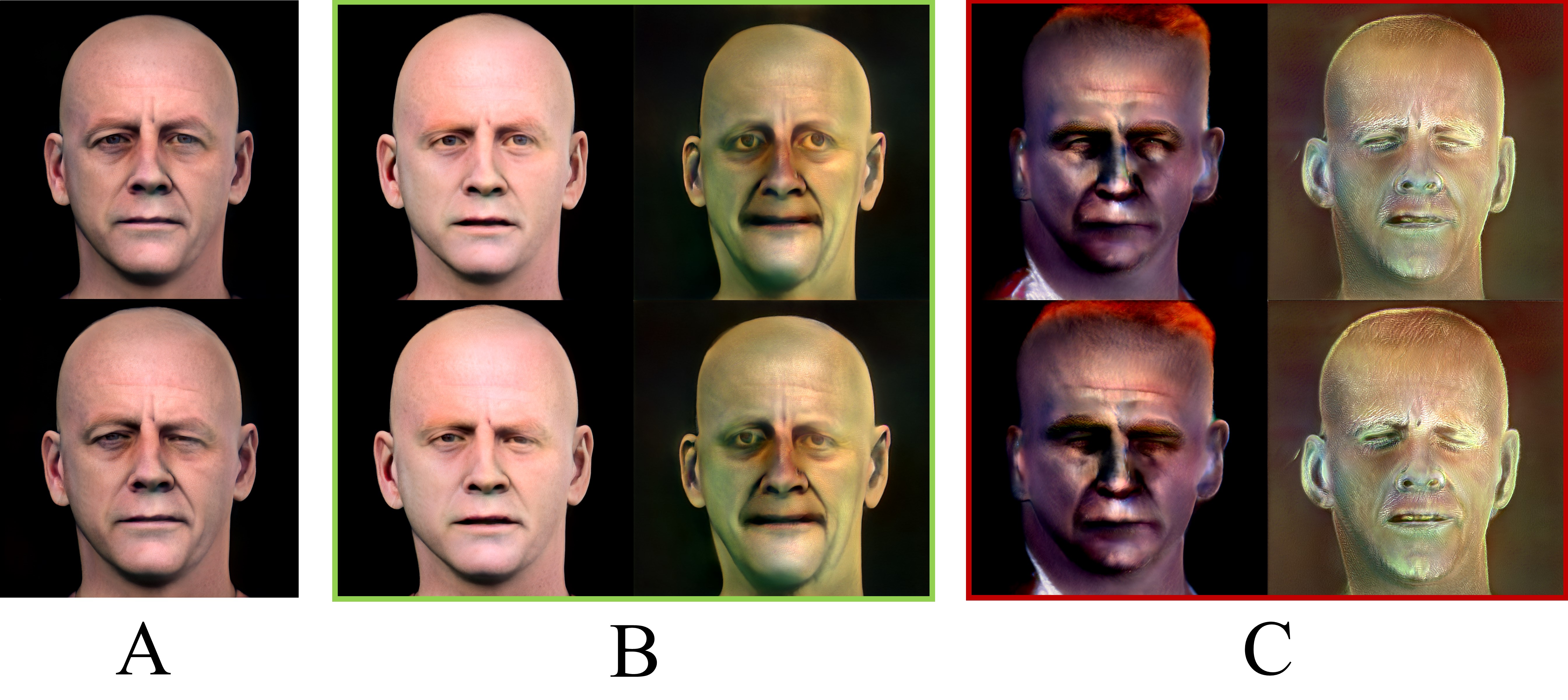}
    \caption{Data augmentation by adding noise to the latent vectors:
    A - the original samples before and after the change $\bb{a}_i, \bb{a}_j$.  B - small-amplitude noise ($\sigma^2=1, \sigma^2=2$) is added to the latent vectors such that the same $\bb{\Delta}$ between the images is preserved with no additional relative changes. C - Too much noise is added ($\sigma^2=4, \sigma^2=5$), making the images differ by more attributes than desired (i.e., the left pair is also differentiated by the amount of hair; the right pair by opening the mouth).}
    \label{fig:data augmentation}
\end{figure}

To avoid over-fitting, we enrich the data by adding random noise 
$\bb{n} \sim \mathcal{N}(0,\,\sigma^2)$ to every latent vector that belongs to the same class during the training phase.\\
\begin{equation}
    \begin{cases*}
    \bb{a}_i \leftarrow \bb{a}_i + \bb{n_a} \\
    \bb{b}_i \leftarrow \bb{b}_i + \bb{n_b}
    \end{cases*}
\end{equation}
Since tiny latent space perturbations in StyleGAN change the semantics in a disentangled manner, adding the same relatively small noise to different latent vectors in $W^+$ preserves semantic changes between them. On the other hand, above a certain magnitude, a noise might transfer the latent vectors to other areas in the latent space that do not maintain the exact change of semantics. Examples shown in Fig. \ref{fig:data augmentation}.\\
Effectively, each attribute's $\bb{\Delta}$ is learned by an unlimited number of samples, instead of only one female and one male.  
An equation that describes a full pipeline of the model would be:
\begin{equation}
    \hat{\bb{b}}_j = D(E(\bb{a}_i+\bb{n_a}, \bb{a}_j+\bb{n_a}), \bb{b}_i+\bb{n_b})+\bb{b}_i.\
\end{equation}

\subsection{Enforcing Linear Space Properties}

To enforce the model learning a relative $\bb{\Delta}$, $\bb{\Delta}_{i,j}$, instead of attribute transfer, we force the  $\Delta$-space to be linear and orthonormal.

We do it during the training phase by decoding the same $\bb{\Delta}_{i,j}$ with all the latent vectors that belong to the same attribute's series $\bb{a}_i, \bb{a}_{i+1},...,\bb{a}_{n-1-(j-i)}$, adding $\alpha \cdot \bb{\Delta}_{i,j}$, and demanding the matching endpoints. To enforce the linear space properties, the parameter $\alpha$ is a scalar set according to the following equation:
\begin{equation}
     D(\bb{a}_k,\alpha \cdot \bb{\Delta}_{i,j})=\hat{\bb{a}}_{k+\alpha\cdot(j-i)}.
\end{equation}

\begin{figure}[]
    \centering
    \includegraphics[width =0.95\columnwidth ]{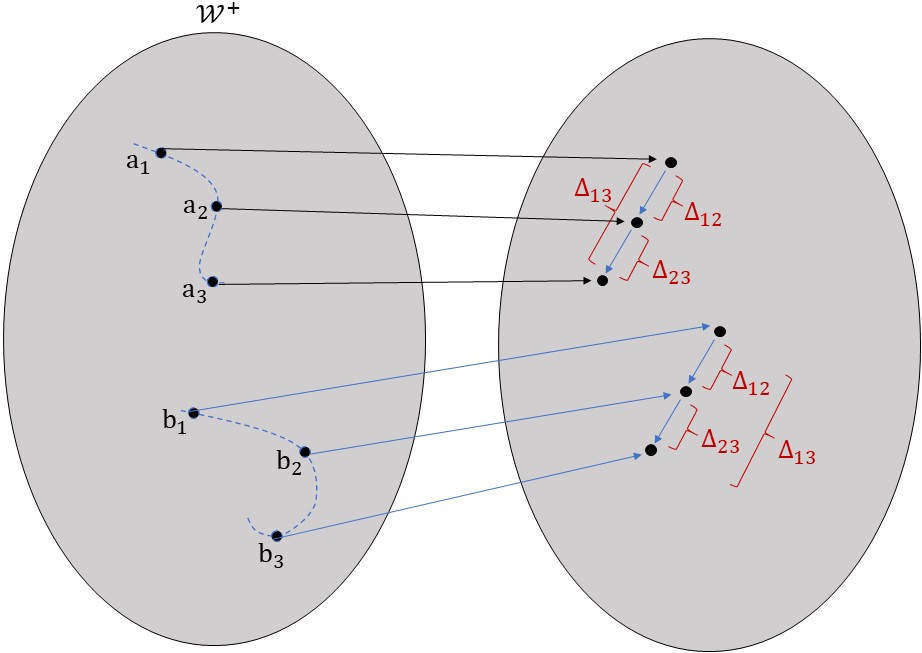}
    \caption{Enforcing linear behavior in the $\Delta$-space: The left ellipse demonstrates the behavior in the GAN's $W^+$ space; the paths of two series of the same attribute changing along a scale of two different classes are shown. The paths are different, nonlinear, and not scaled. The right ellipse demonstrates the transformation enforced in the $\Delta$-space: the difference between two consecutive images, as represented in the $\Delta$-space, is scaled and oriented similarly, and is uniform among the different classes. Also, the addition of consecutive $\bb{\Delta}$-s constructs the cumulative $\bb{\Delta}$.}
    \label{fig:demand linearity}
\end{figure}

An illustration of the linearity constraint is shown in Fig. \ref{fig:demand linearity}.

Our model learns the $\bb{\Delta}$-s between the images with two biases: (1) A bias caused by the distribution difference between the synthetic samples and the real images. (2) A bias caused by the imperfect projection method in terms of reconstruction and editability. To mitigate these issues, we validate the editability of projected real images and, even more importantly, the editability of random images the GAN synthesized, since the latent space semantics are best reflected on them.
\label{Enforcing linear space properties}

\section{Experiments}
\subsection{Data Analysis}
\begin{figure}[]
    \centering
    \subfloat[Real face images VS synthetic data in the latent space]{\includegraphics[width=\columnwidth ]{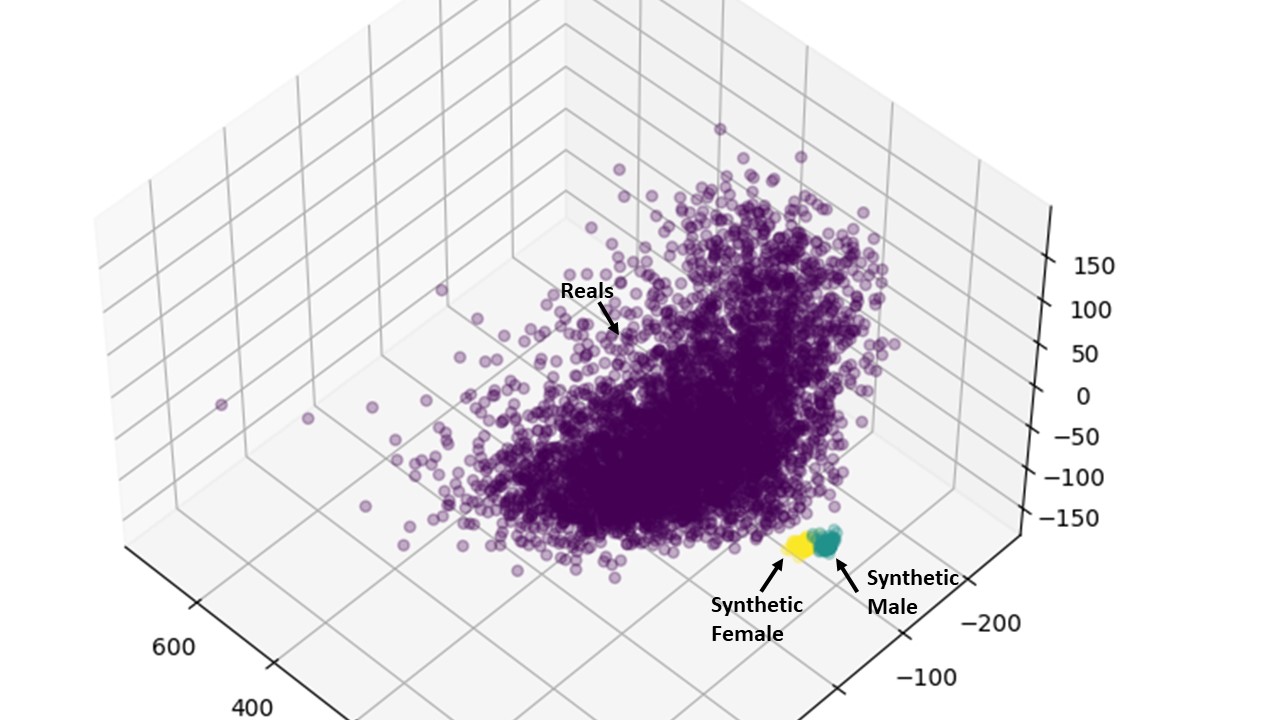}}\\
    \subfloat[Deltas of latent vectors, relating the same expression change: Real faces VS synthetic space]{\includegraphics[width=\columnwidth ]{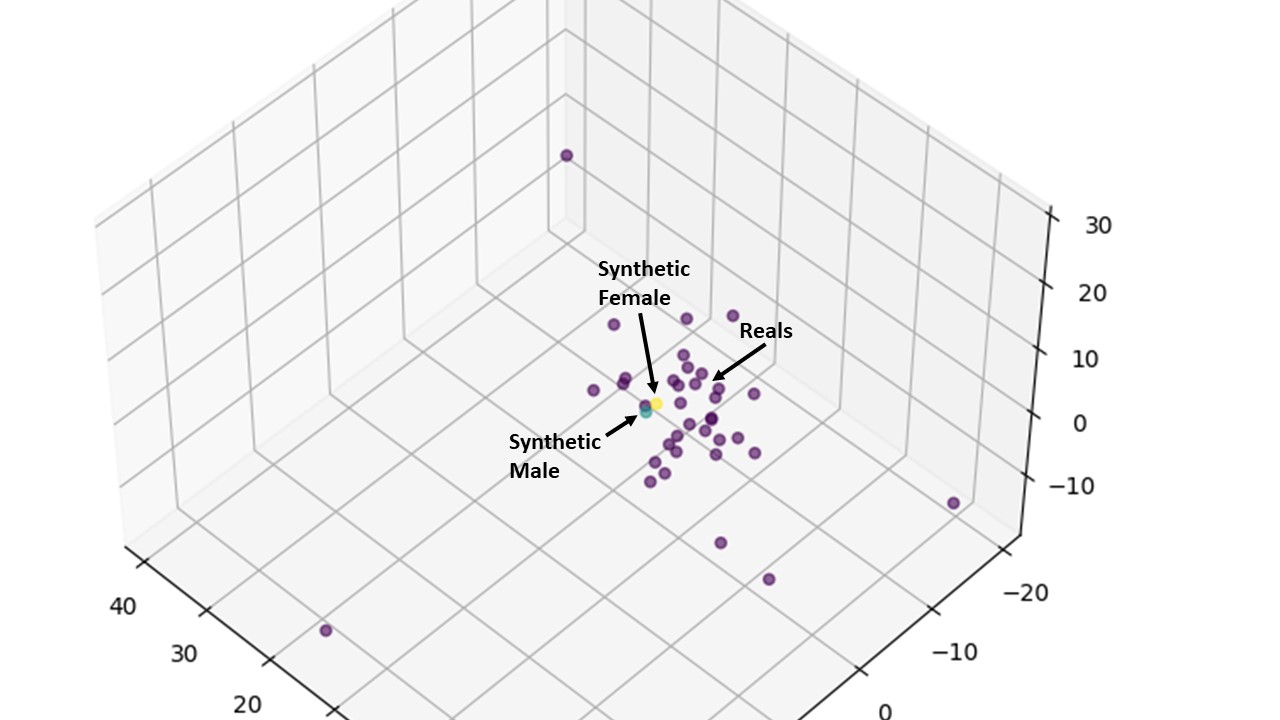}}
    
    \caption{PCA dimension reduction of projected synthetic data in comparison to projected real-world data, as well as comparison between the latent vector differences that these groups define.
}
    \label{fig:distribution-pca}
\end{figure}

First, to observe the relation between the synthetic data and images of real faces, we plot a low dimension distribution of the projection for both of them. The projection shows that the synthetic data is separated from the distribution of the real-world images. Nevertheless, linear interpolation of the differences corresponding to the same attribute change fall inside the same distribution, thus allowing studying from synthetic data for later inferring over real-world samples (Fig. \ref{fig:distribution-pca}).

\subsection{Comparison to Linear Paths}
We show a series of comparisons to the linear paths edits that matches the state-of-the-art projection of ReStyle \cite{alaluf2021restyle} based on PSP \cite{richardson2021encoding}. 
We compare editing different attributes to show the comprehensive ability of the technique: different expressions that combine several changes (e.g., emotions like happy or sad), and expressions that change only a single attribute (e.g., winking). We also test our model over changes in pose and lighting. Our results outperform state-of-the-art in terms of FID score (realism of the edited outputs) in Table \ref{tab:fid score}, and in terms of cosine similarity between the output images' features along with sequences of editing (the edited outputs reliably preserve the identity) in Table \ref{tab: cosine similarity}.

\begin{figure*}[]
\centering
\subfloat[Expression change with only one detail change: Closing eyes.]{\includegraphics[width = 0.65\columnwidth]{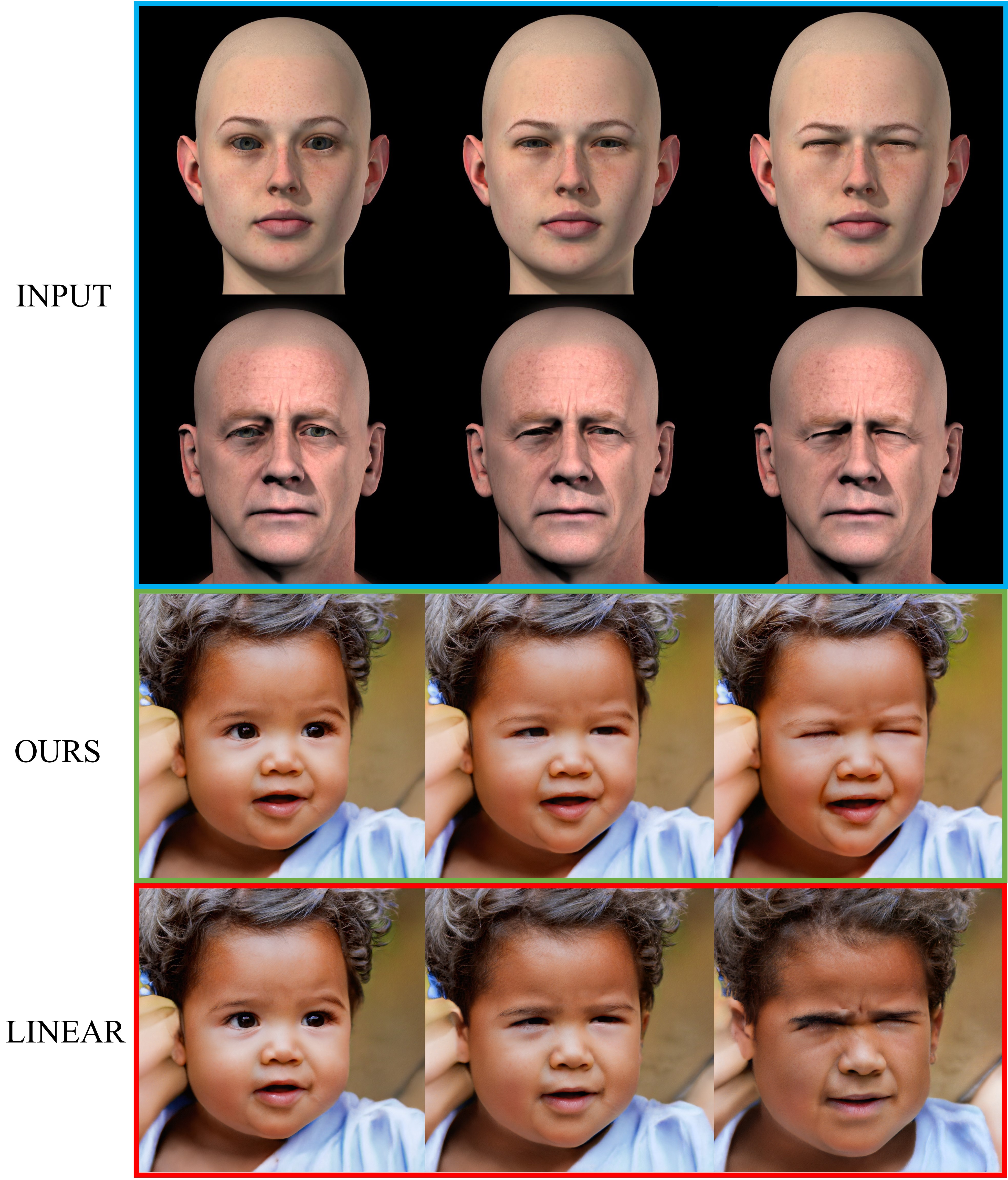}}\hspace{1.45em}
\subfloat[Expression change with several details together: Being happy.]{\includegraphics[width = 0.65\columnwidth]{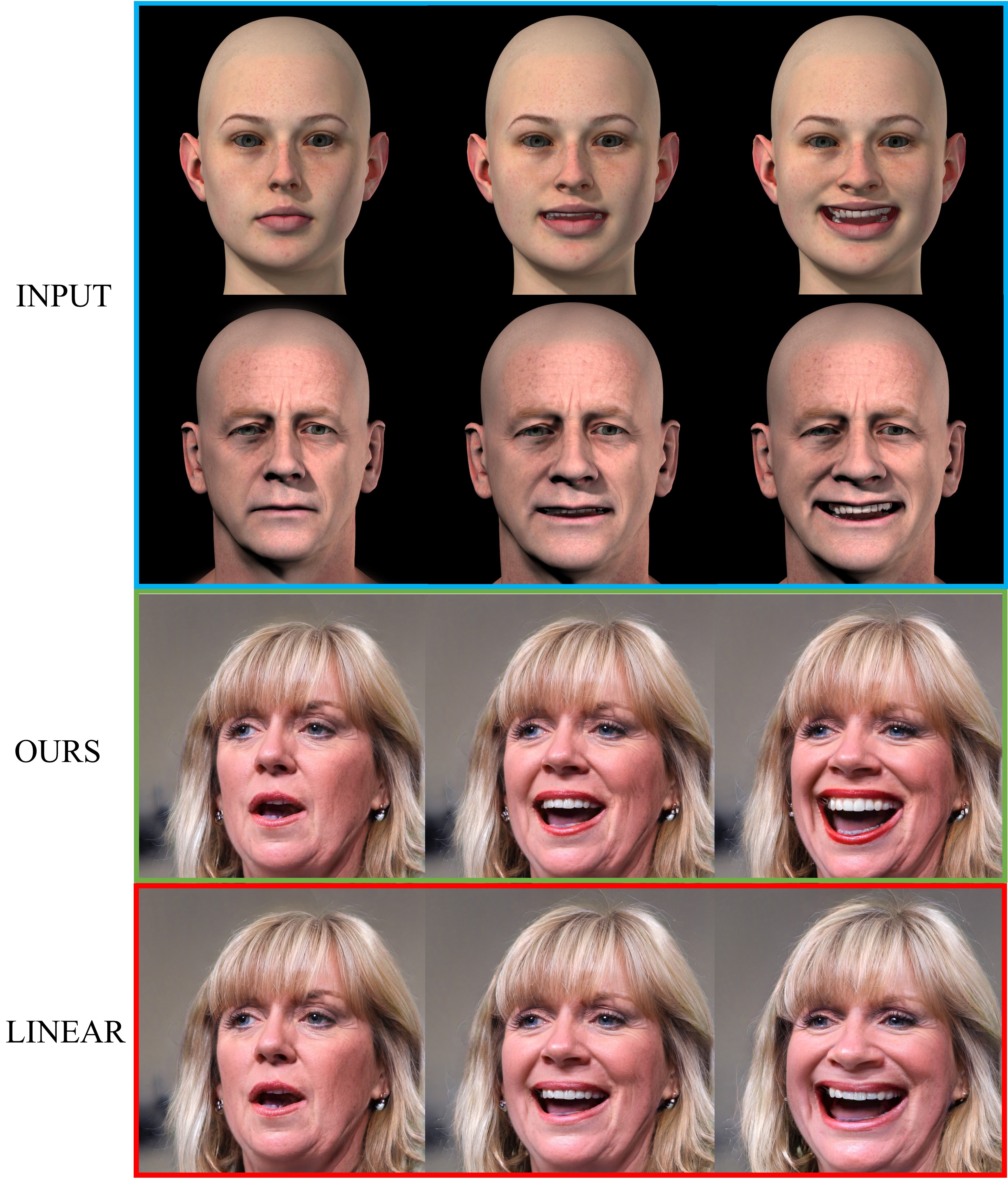}}\hspace{1.45em}
\subfloat[Expression change with several details together: Being sad.]{\includegraphics[width = 0.65\columnwidth]{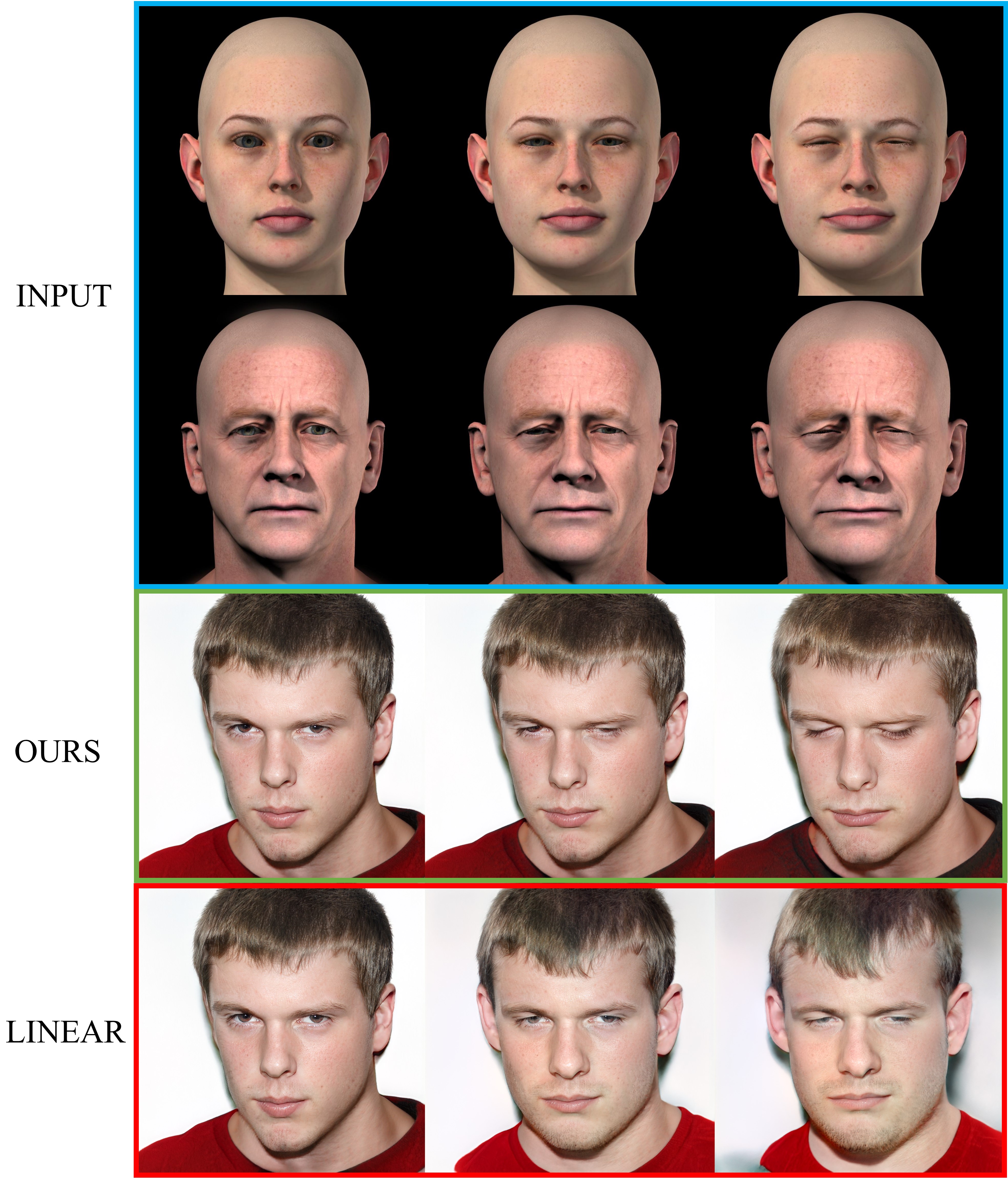}}\\
\subfloat[Pose change: Head right.]{\includegraphics[width = 0.65\columnwidth]{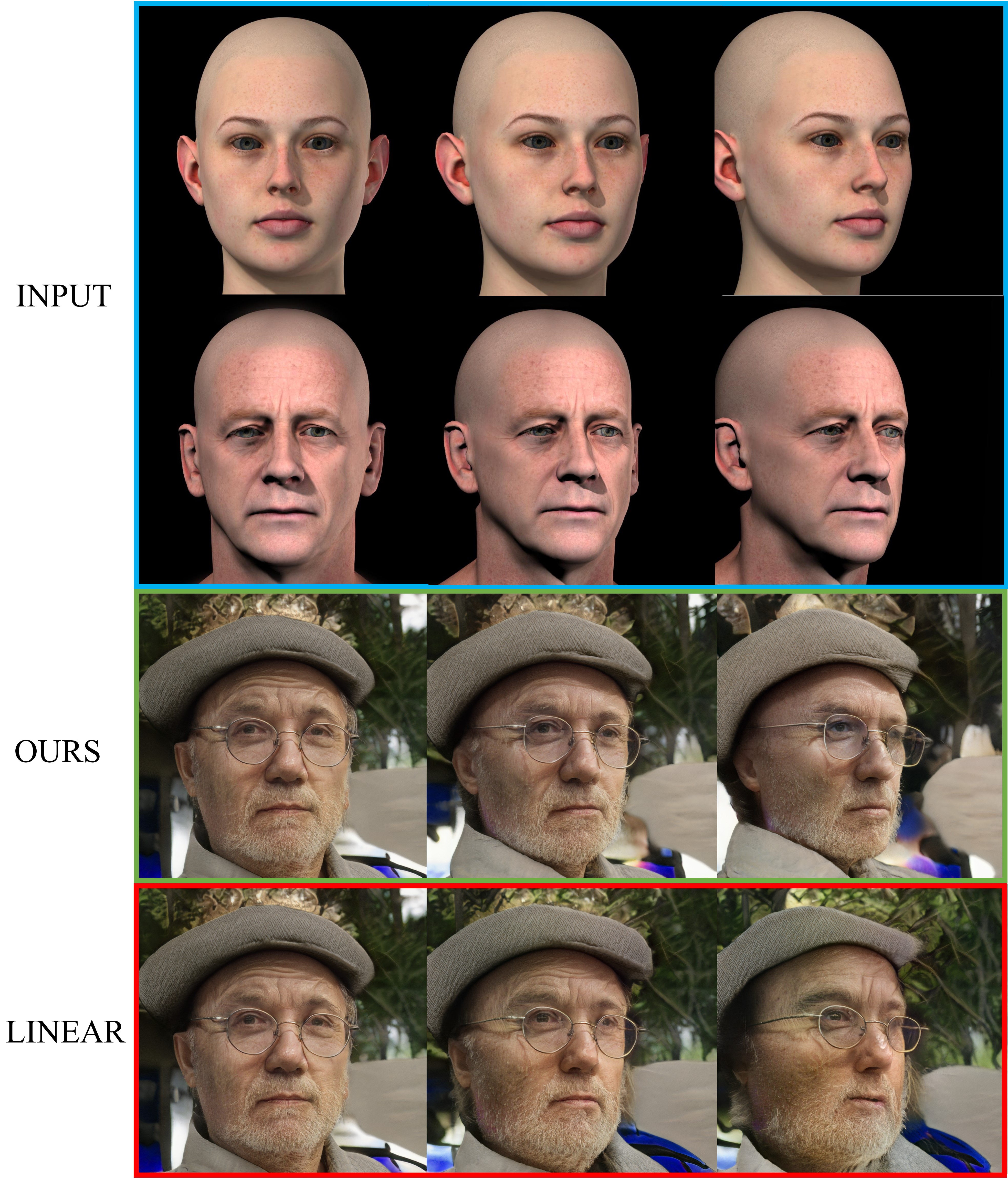}}\hspace{1.45em}
\subfloat[Pose change: Head back.]{\includegraphics[width = 0.65\columnwidth]{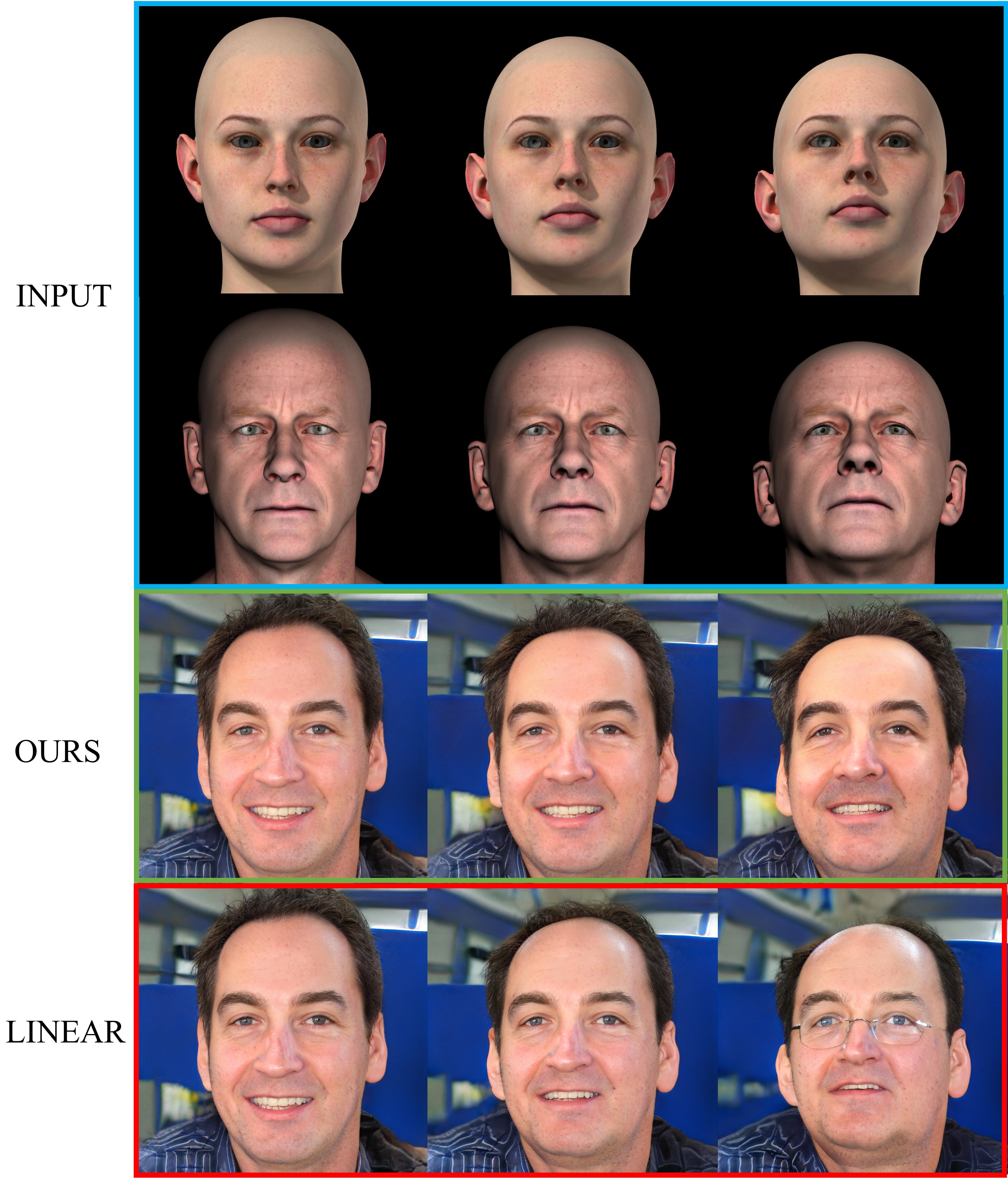}}\hspace{1.45em}
\subfloat[Light change: lighting is approaching from left to right.]{\includegraphics[width = 0.65\columnwidth]{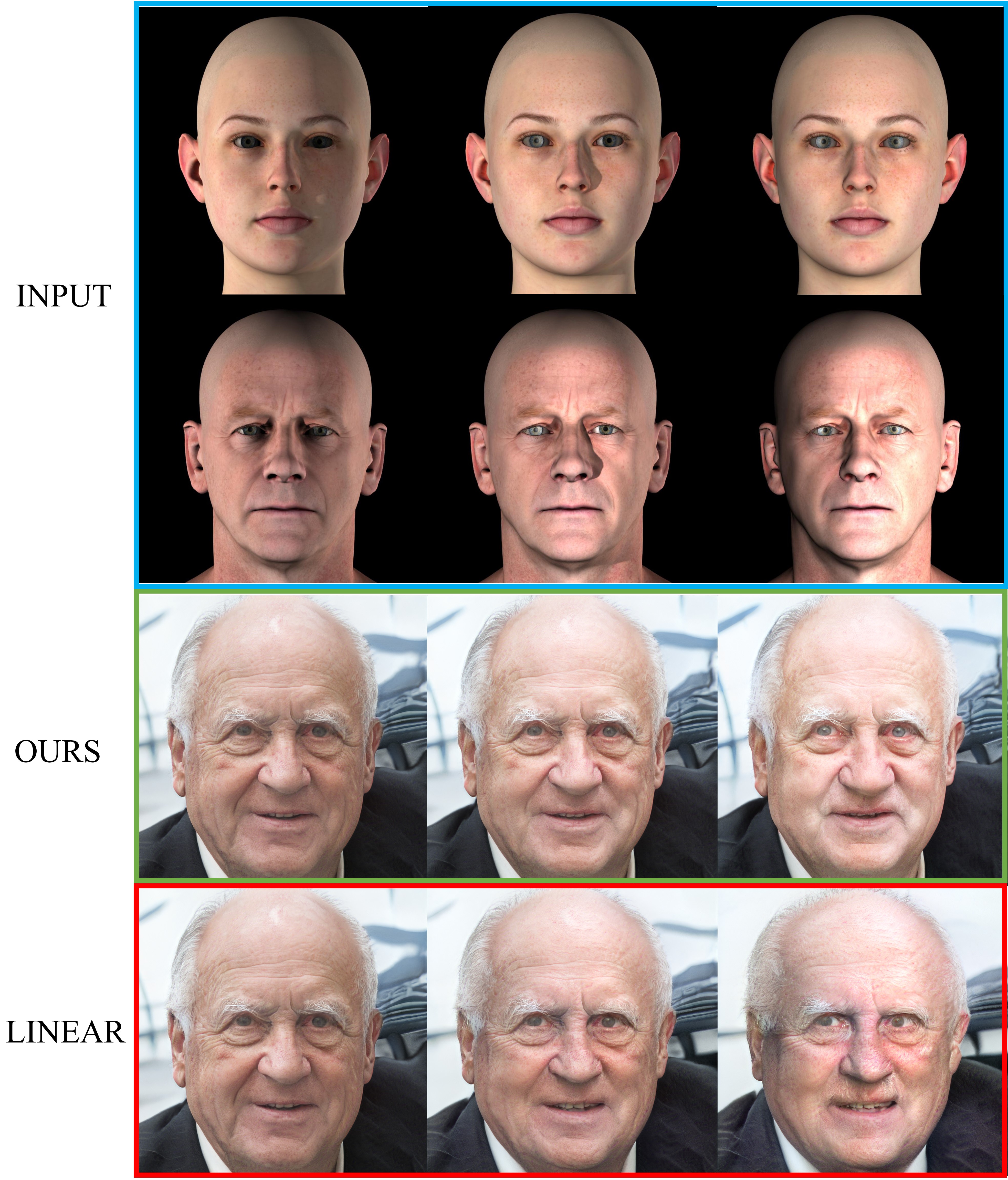}}
\caption{Demonstration of our model's results comparing to the linear path that defines the semantic difference in StyleGAN2.  The first two rows are the input to the model, where the 3rd row corresponds to our model's results, and the last row corresponds to linear editing. The different $\bb{\Delta}$-s corresponds to a single detail, several details, pose or lighting changes.}
\label{figure:experiments random latents}
\end{figure*}

Above each comparison, we show the desired change defined by three images from two synthetic classes in order to demonstrate the input to the model (Fig. \ref{figure:experiments random latents}). The comparison is performed between random latent vectors from the GAN's $W^+$ distribution, emphasizing that our model learns the latent space specific semantics despite the different biases mentioned in subsection \ref{Enforcing linear space properties}.

\begin{figure}[]
\centering
\subfloat[Real people images expression change]{\includegraphics[width = 0.4062625\columnwidth]{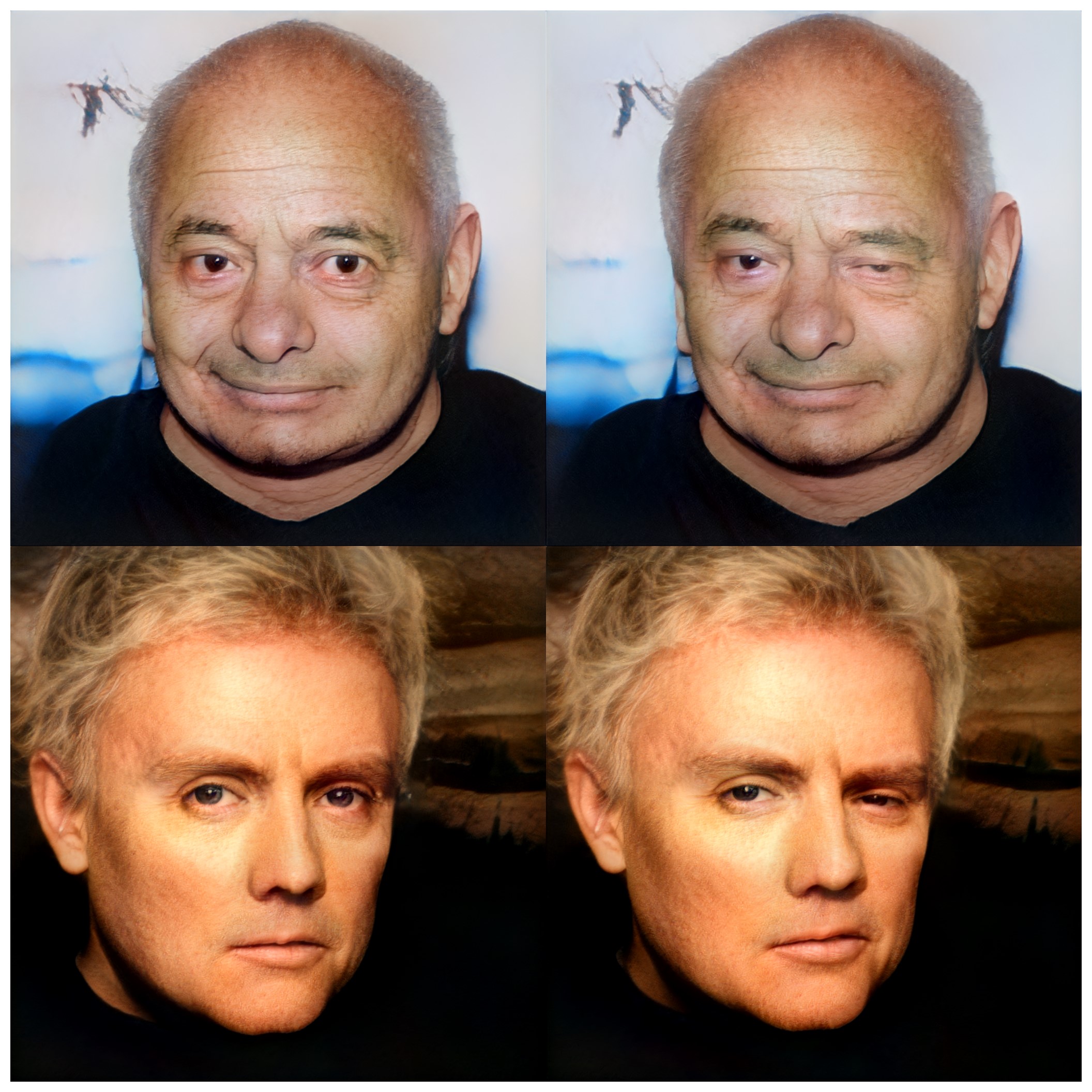}}\\
\subfloat[Real people images pose change]{\includegraphics[width = 0.4062625\columnwidth]{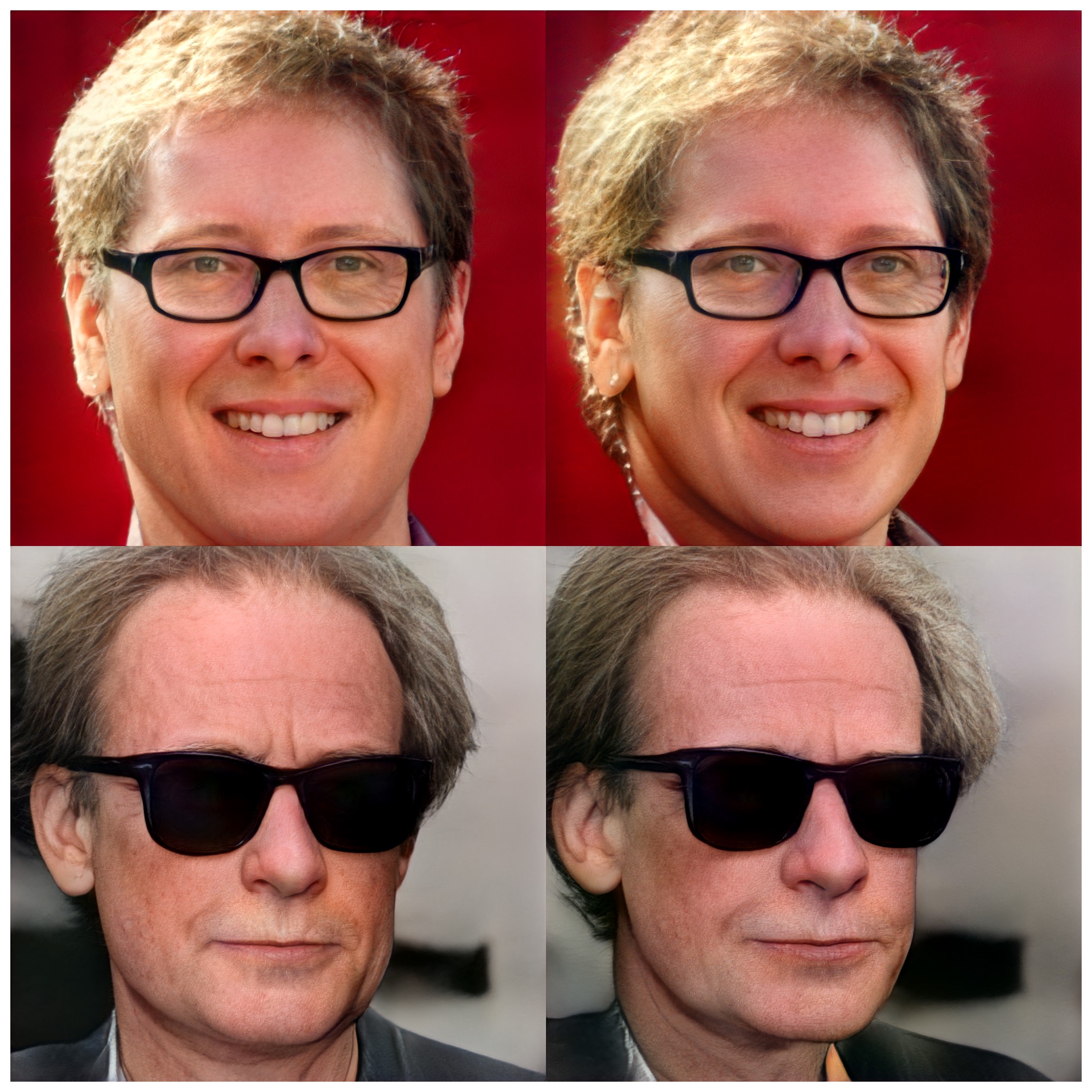}}\\
\subfloat[Real people images lighting change]{\includegraphics[width = 0.4062625\columnwidth]{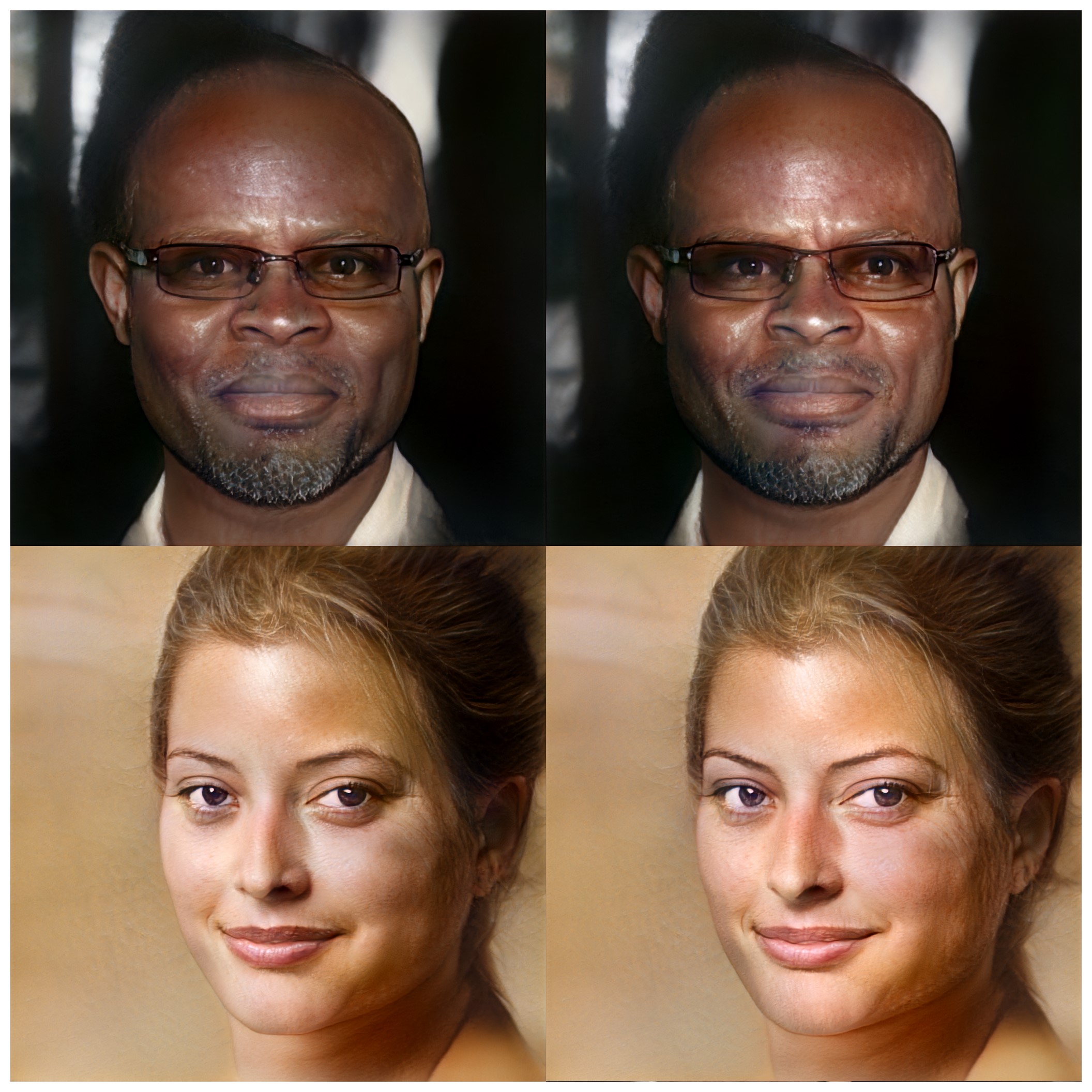}}
\caption{Demonstration of our model's results, applying $\bb{\Delta}$-s to real people images, over expression, pose and lighting.}
\label{figure:experiments-real-people}
\end{figure}

We then show editing of different attributes on images of real people projected to $\bb{W^+}$ vectors by the same manner we had projected our synthetic models (Fig. \ref{figure:experiments-real-people}).

\begin{figure}[]
\centering
\subfloat{\includegraphics[width=0.5\columnwidth]{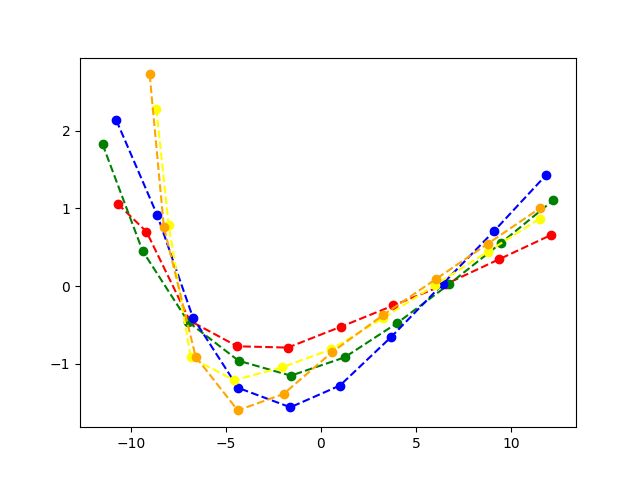}}
\subfloat{\includegraphics[width=0.5\columnwidth]{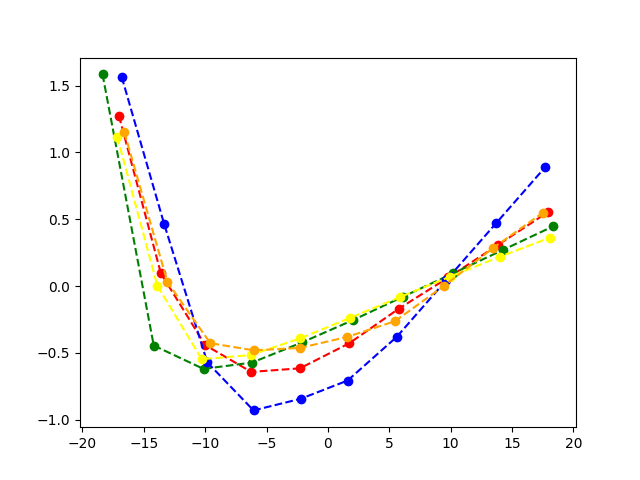}}\\
\subfloat{\includegraphics[width=0.5\columnwidth]{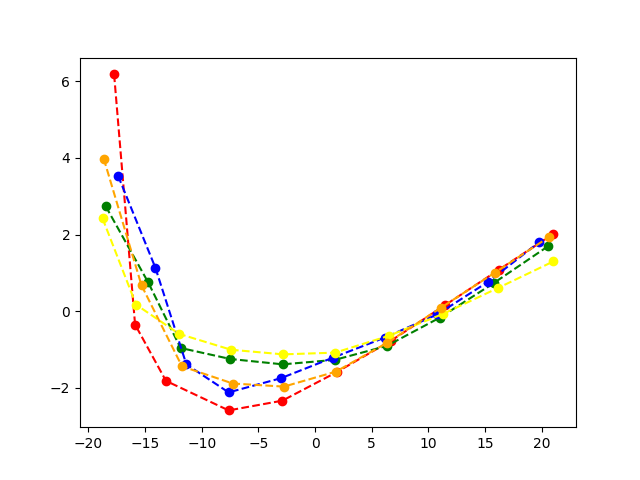}}

\caption{PCA dimension reduction of edit paths in the latent space, consisting of $\bb{\Delta}$-s of expression, pose and light change. Each graph corresponds to 5 different samples edited with the same $\bb{\Delta}$. All paths are nonlinear and change as a function of the input image.}
\label{figure:latent-space-paths-pca}
\end{figure}

To assure the nonlinearity of the $\bb{\Delta}$-s, we present graphs representing the nonlinear paths in the GAN's latent space that correspond to the edits for each attribute family (expression, pose, lighting) as shown in Fig. \ref{figure:latent-space-paths-pca}.

\subsection{Unsupervised Precise Pose Control}
\begin{figure}
    \setlength{\tempwidth}{.3\linewidth}
    \settoheight{\tempheight}{\includegraphics[width=\tempwidth]{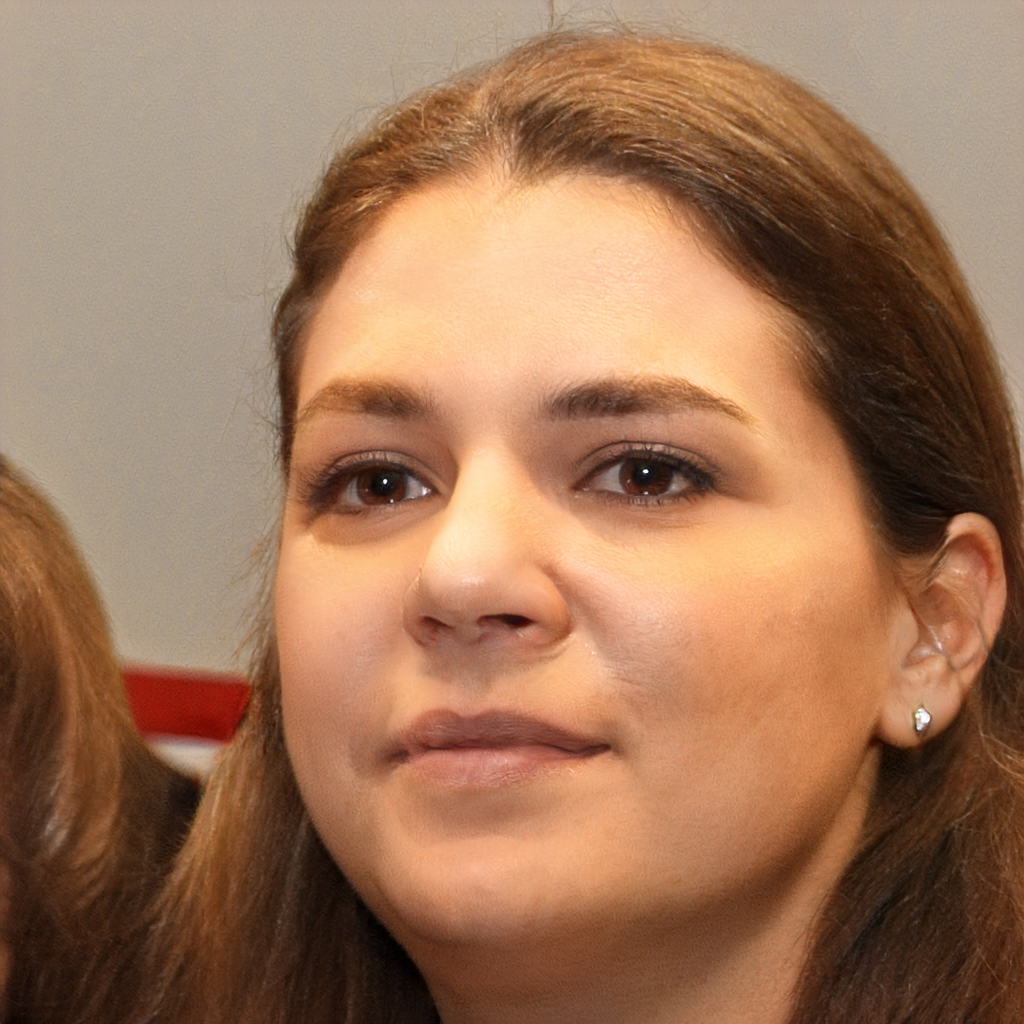}}%
    \centering
    \hspace{\baselineskip}
    \columnname{$\bb{\Delta}=0\degree$}\hfill
    \columnname{$\bb{\Delta}=10\degree$}\hfill
    \columnname{$\bb{\Delta}=20\degree$}
    \subfloat{\includegraphics[width=0.203125\columnwidth]{images/pose_control_images/latent_pose_woman_0.png}}
    \subfloat{\includegraphics[width=0.203125\columnwidth]{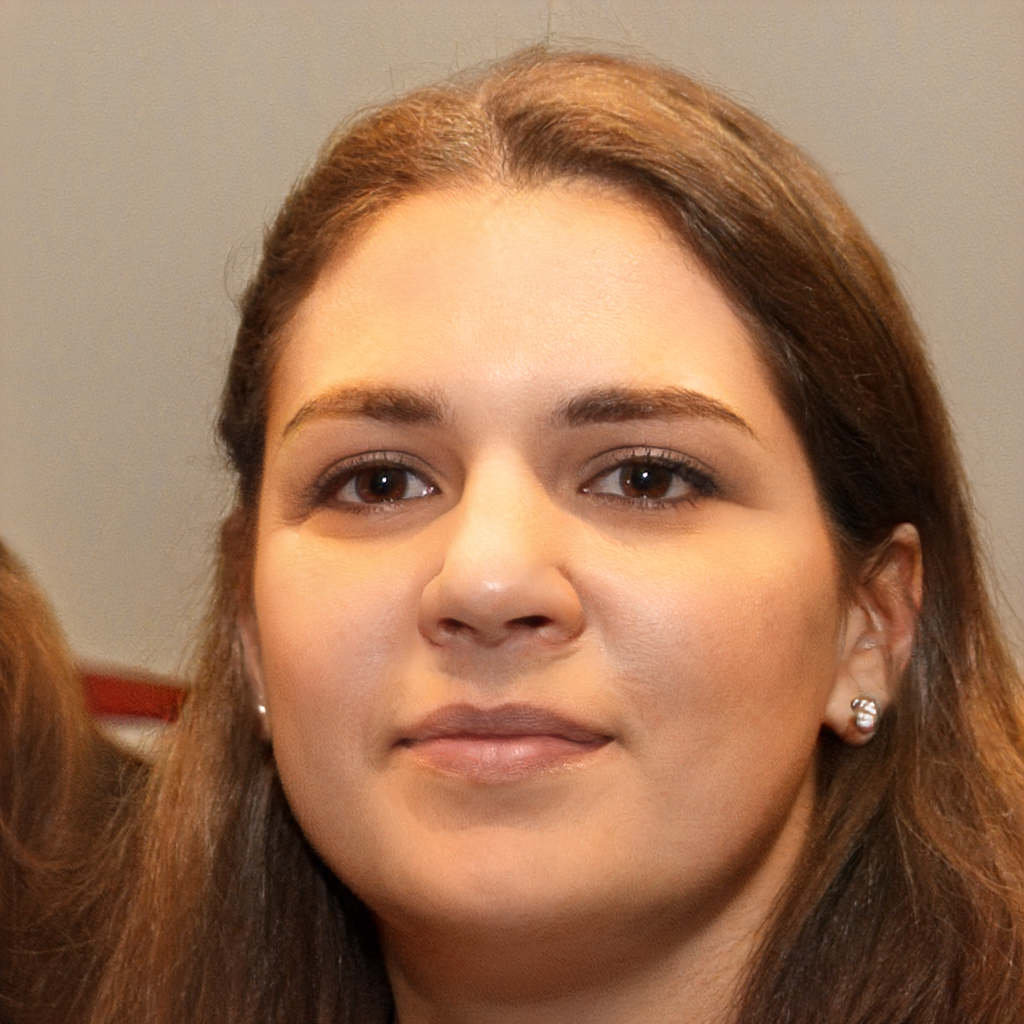}}
    \subfloat{\includegraphics[width=0.203125\columnwidth]{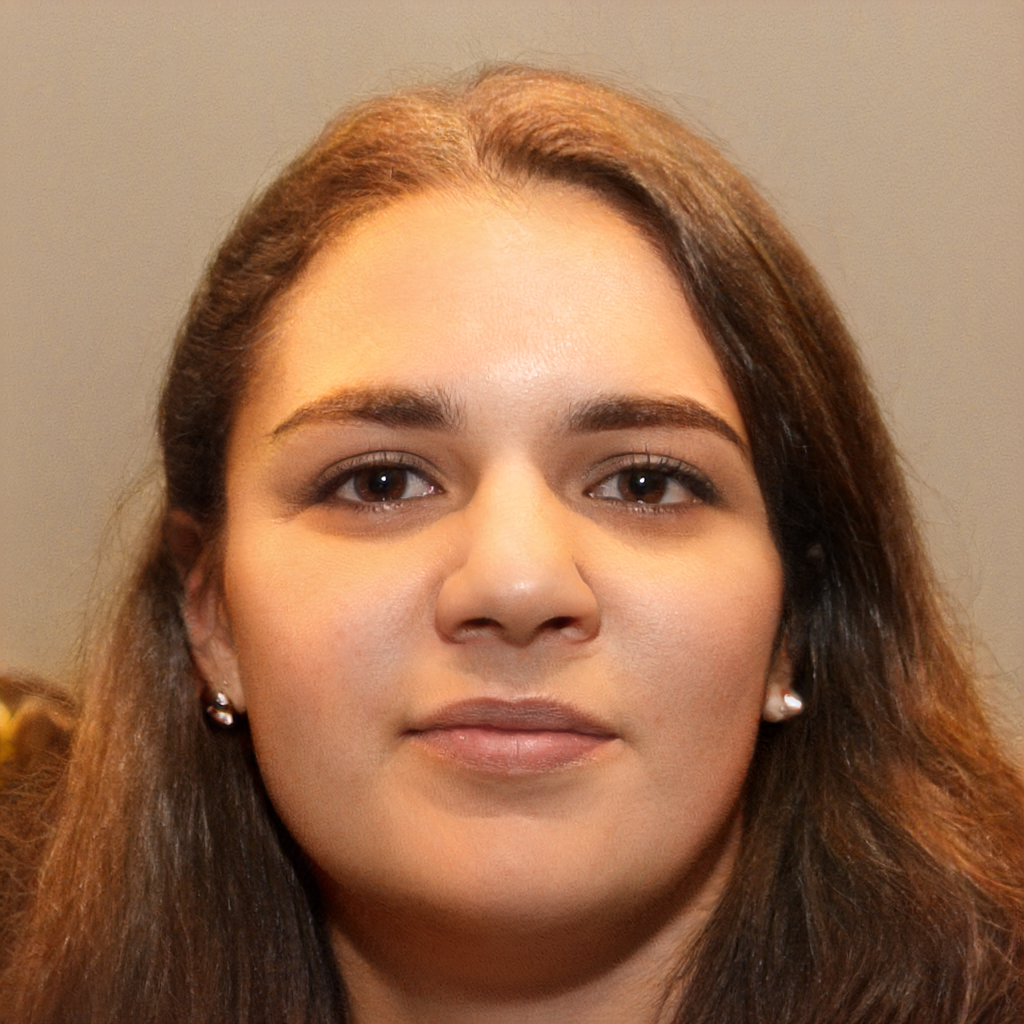}}\\[-2.5ex]
    
    \subfloat{\includegraphics[width=0.203125\columnwidth]{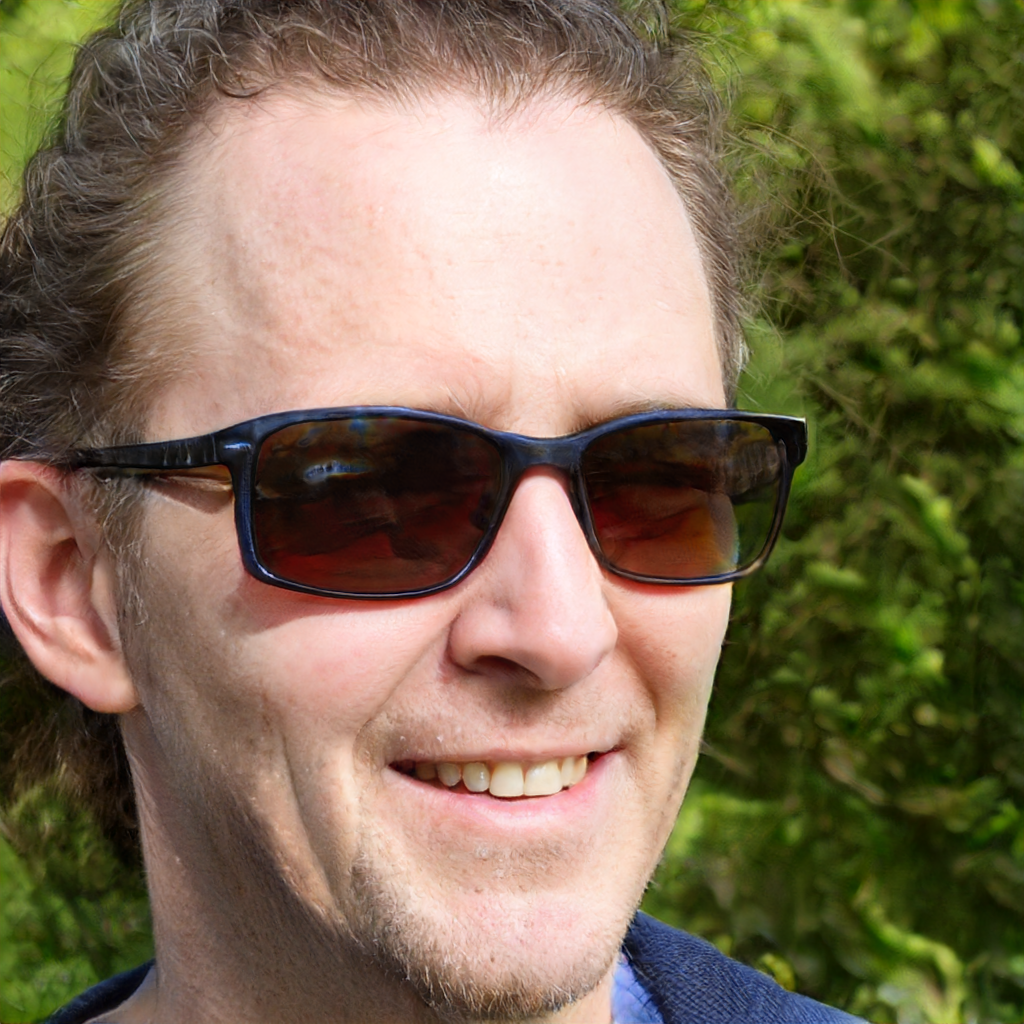}}
    \subfloat{\includegraphics[width=0.203125\columnwidth]{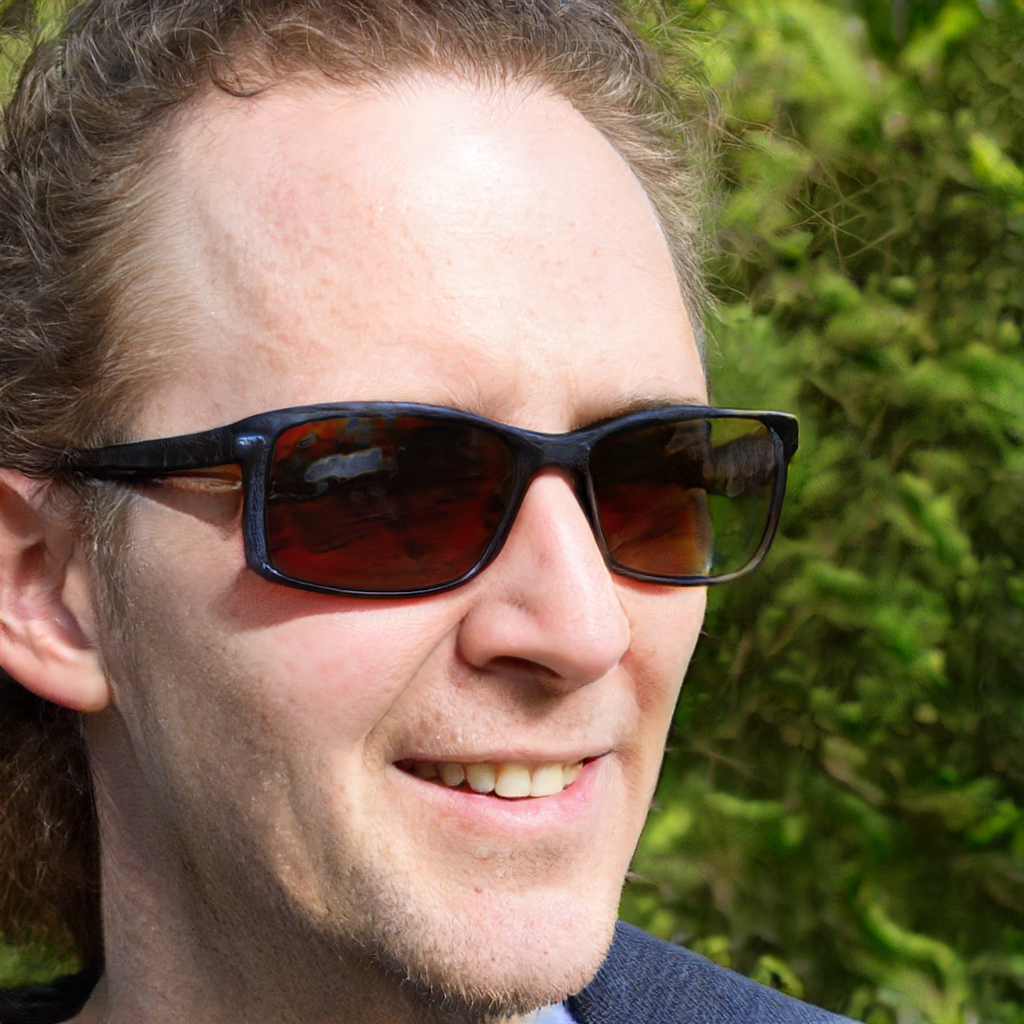}}
    \subfloat{\includegraphics[width=0.203125\columnwidth]{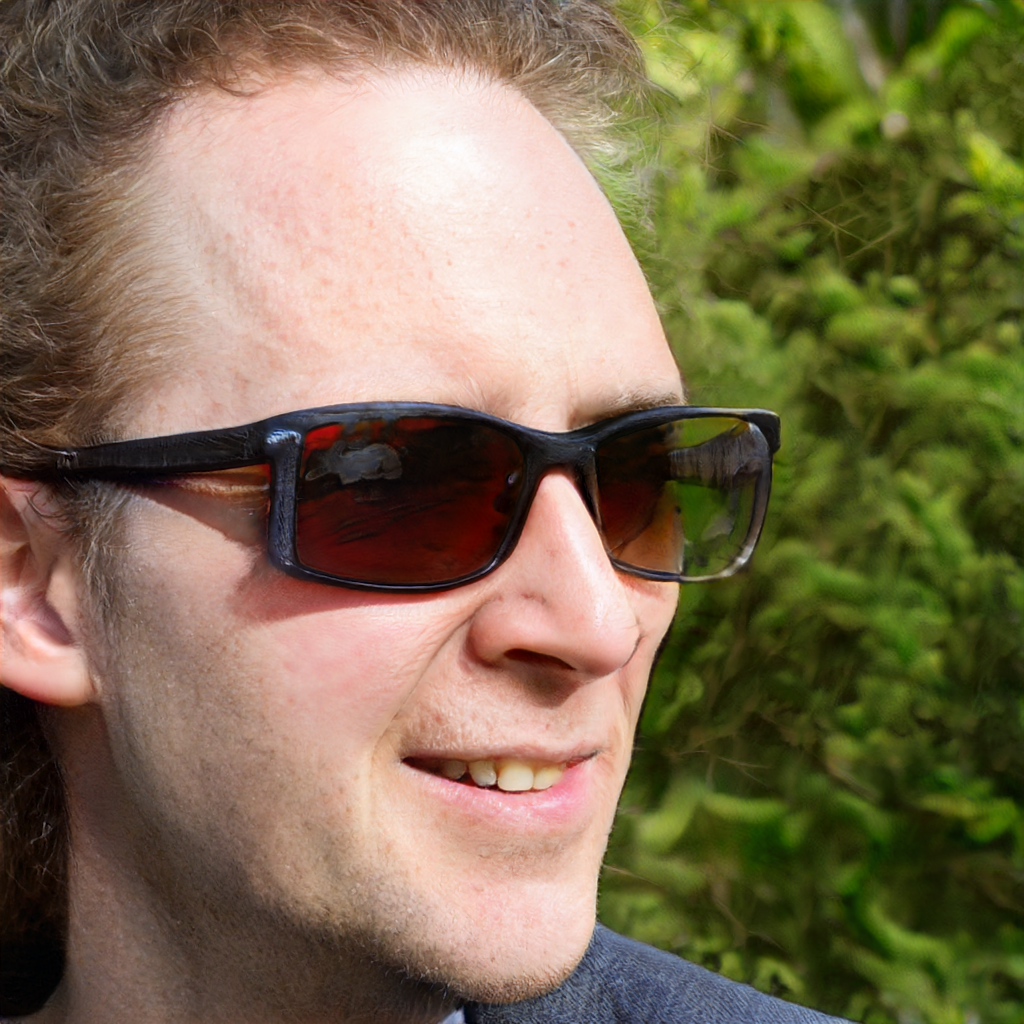}}\\[-2.5ex]
    
    \subfloat{\includegraphics[width=0.203125\columnwidth]{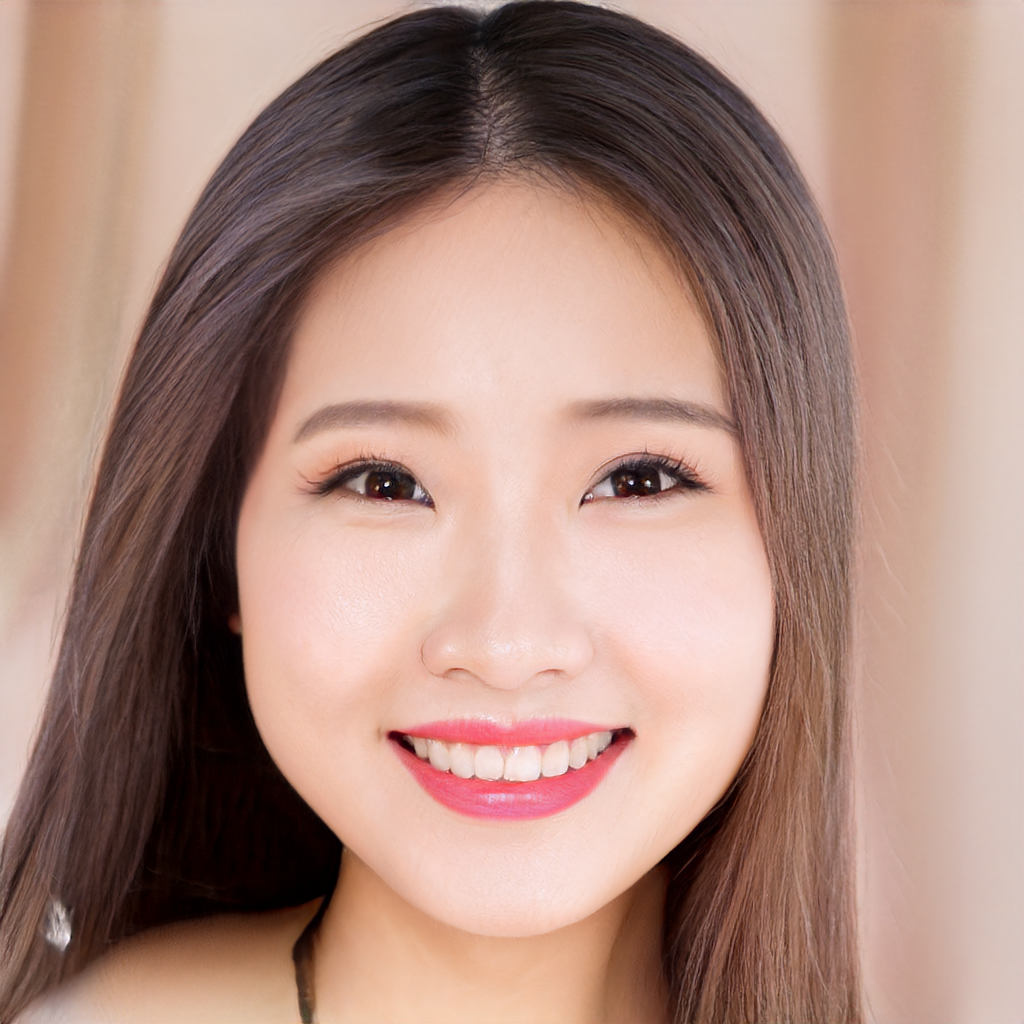}}
    \subfloat{\includegraphics[width=0.203125\columnwidth]{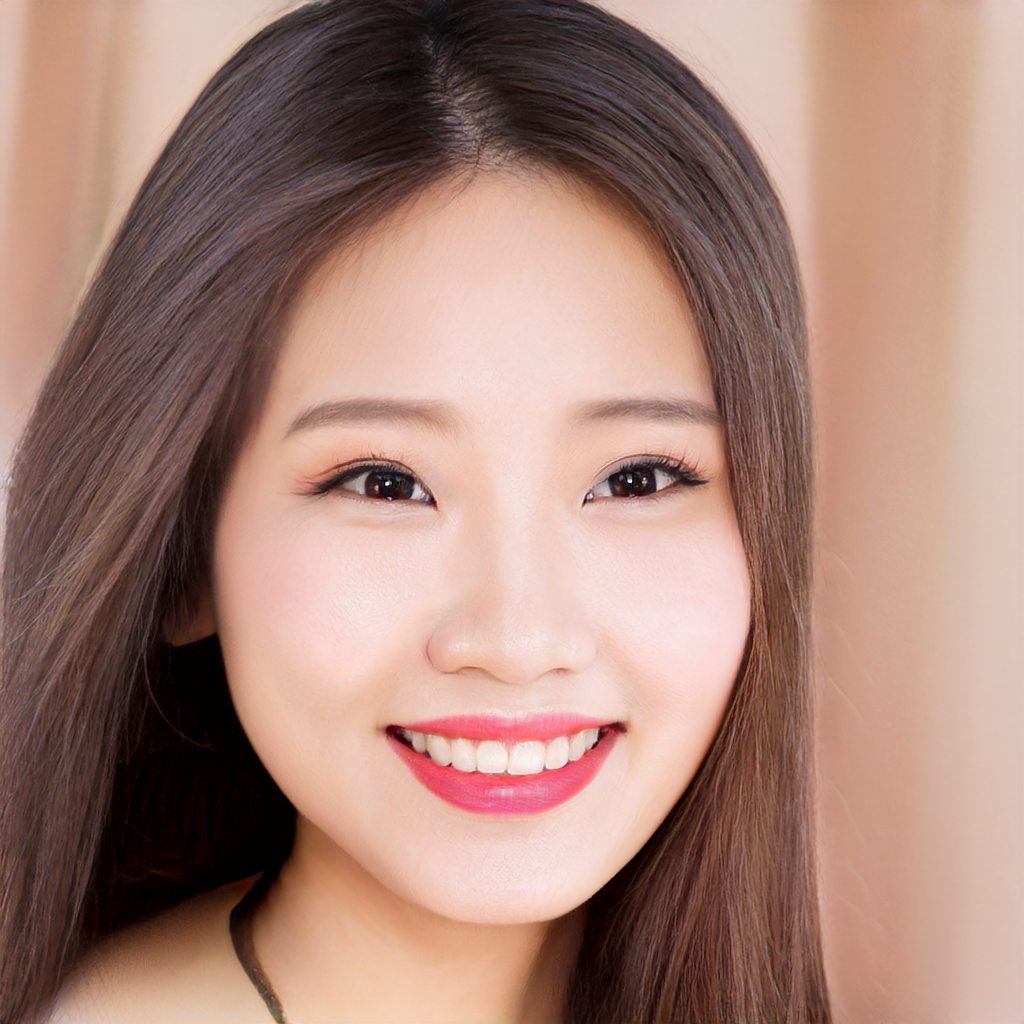}}
    \subfloat{\includegraphics[width=0.203125\columnwidth]{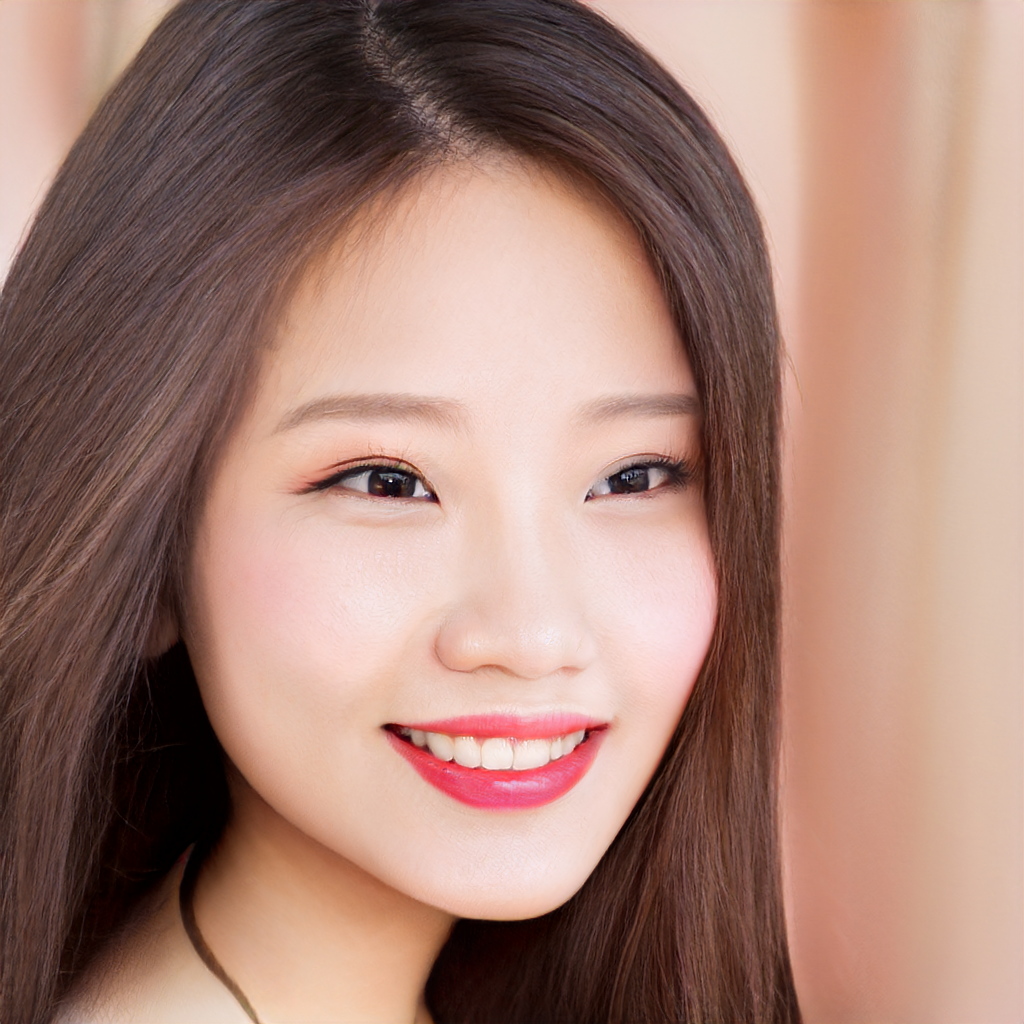}}\\
    
    \caption{Generation of images with precise $\bb{\Delta}$-s of head pose angles, without any pre-trained GAN supervision.}
    \label{fig:pose control synthesized}
\end{figure}

\begin{figure}
    \setlength{\tempwidth}{.3\linewidth}
    \settoheight{\tempheight}{\includegraphics[width=\tempwidth]{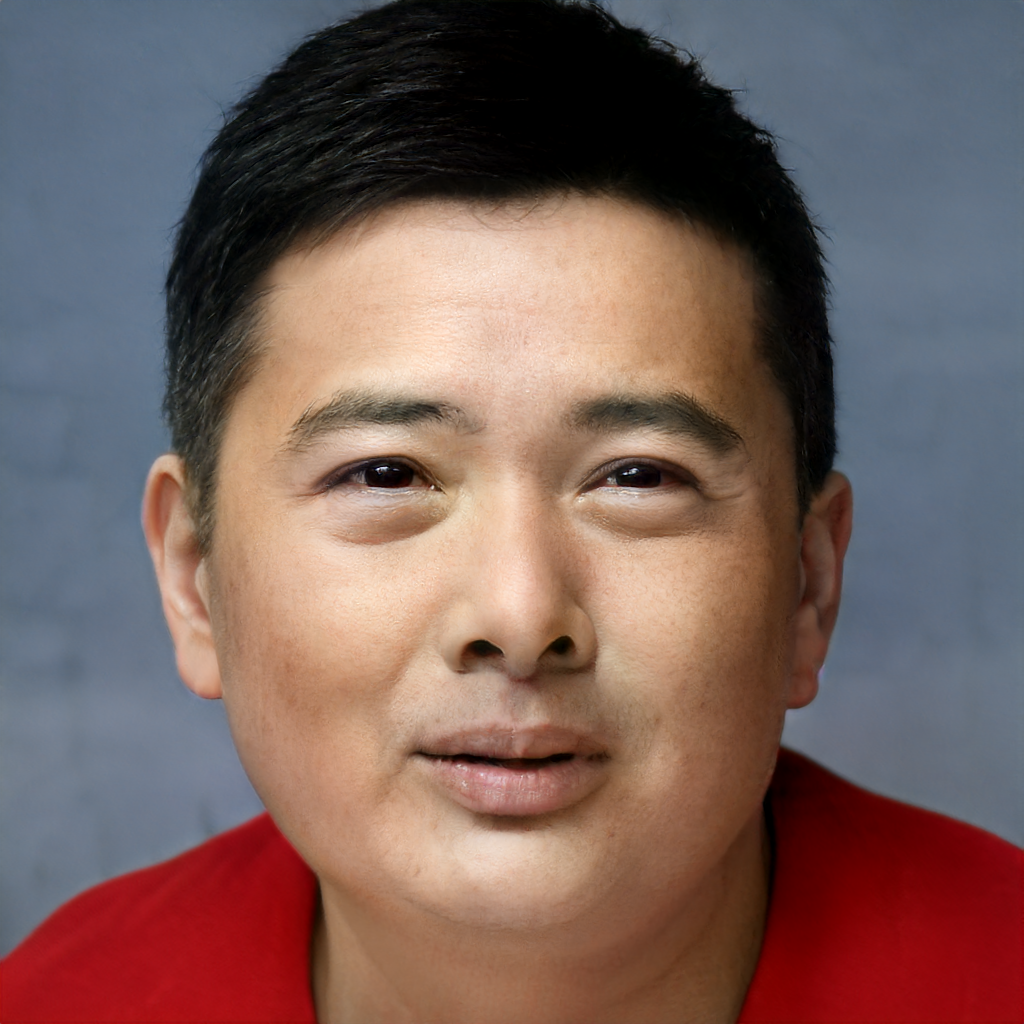}}%
    \centering
    \hspace{\baselineskip}
    \columnname{$\bb{\Delta}=0\degree$}\hfill
    \columnname{$\bb{\Delta}=10\degree$}\hfill
    \columnname{$\bb{\Delta}=20\degree$}
    \subfloat{\includegraphics[width=0.203125\columnwidth]{images/pose_control_images/real_pose_chinese_0.png}}
    \subfloat{\includegraphics[width=0.203125\columnwidth]{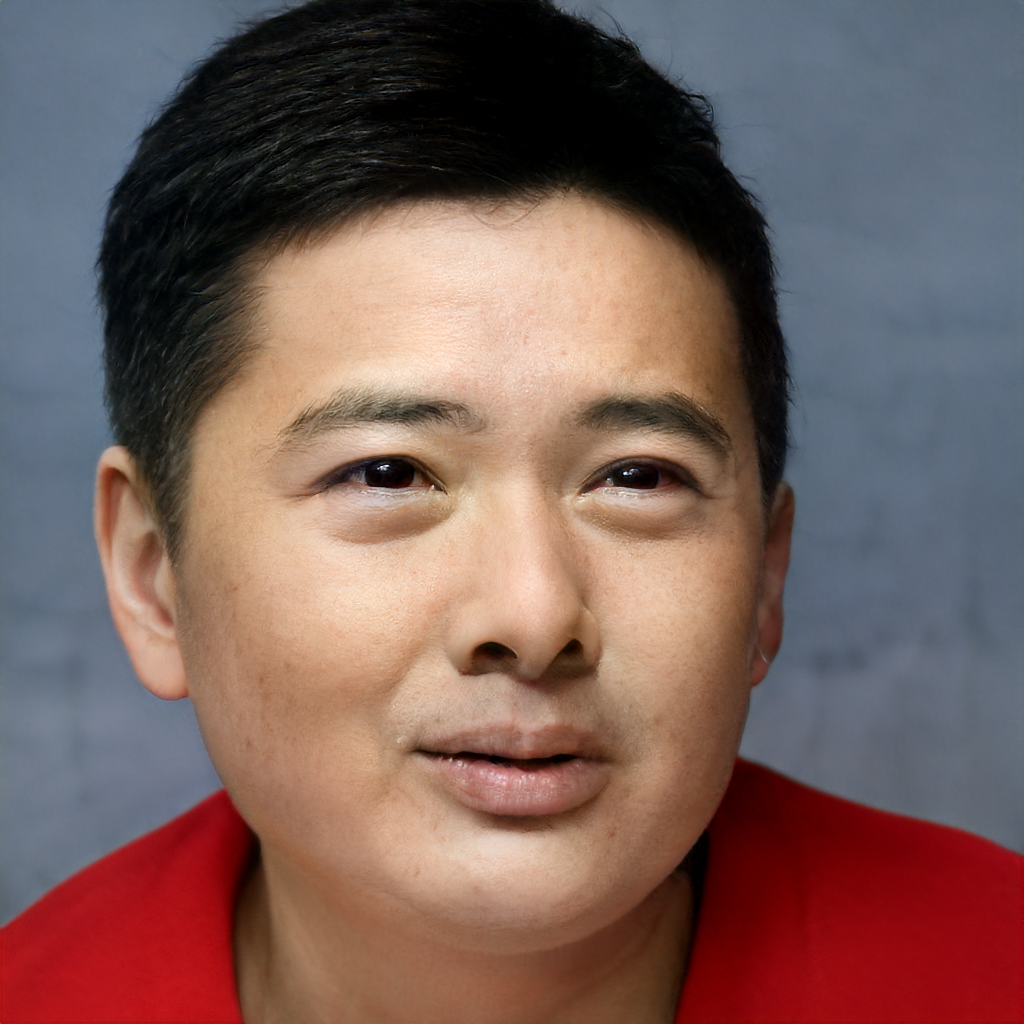}}
    \subfloat{\includegraphics[width=0.203125\columnwidth]{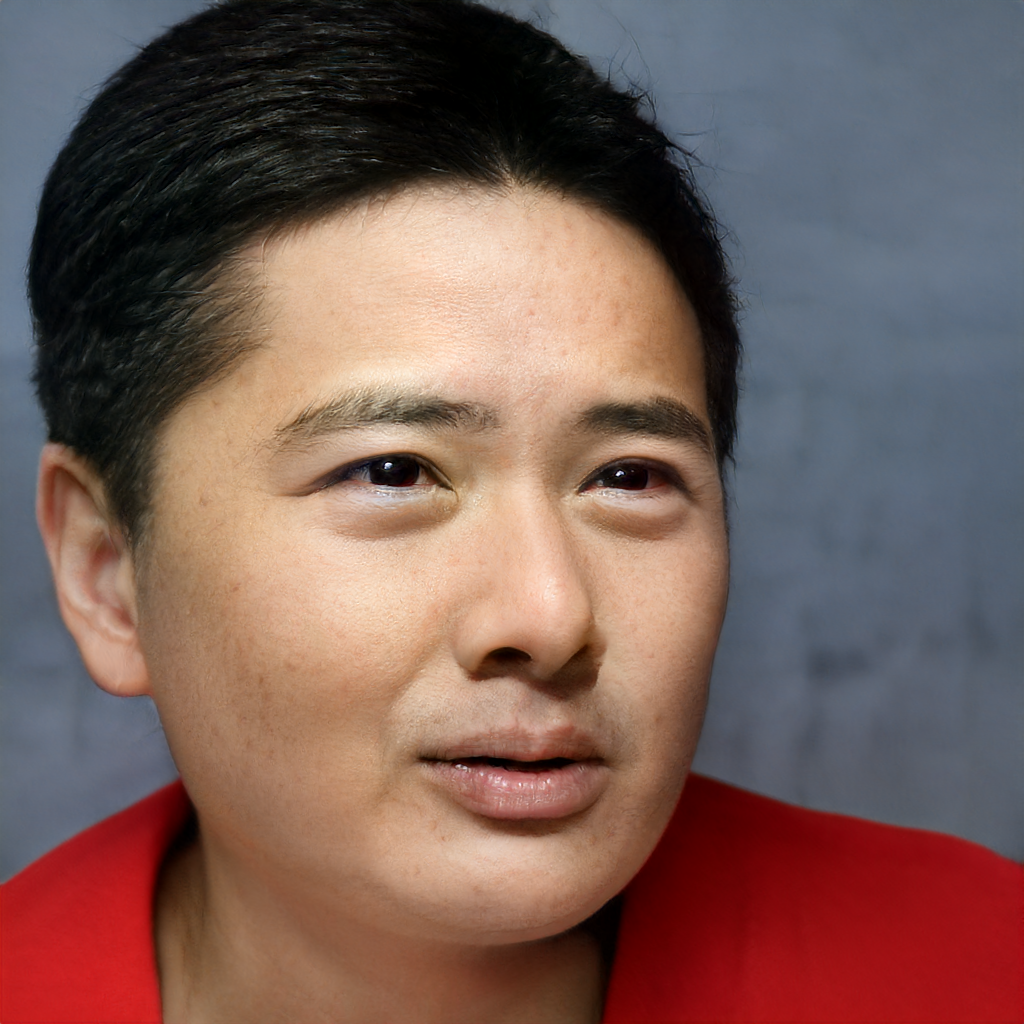}}\\[-2.5ex]
    
    \subfloat{\includegraphics[width=0.203125\columnwidth]{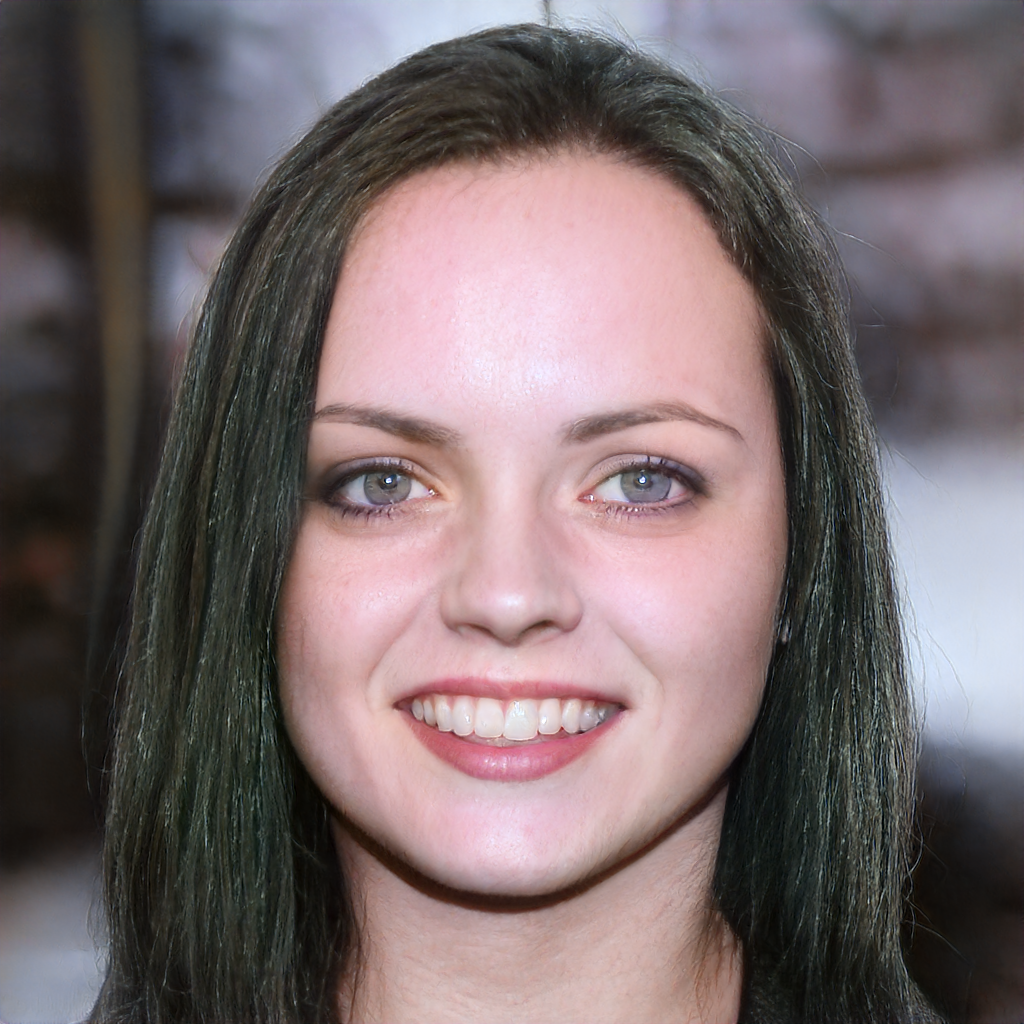}}
    \subfloat{\includegraphics[width=0.203125\columnwidth]{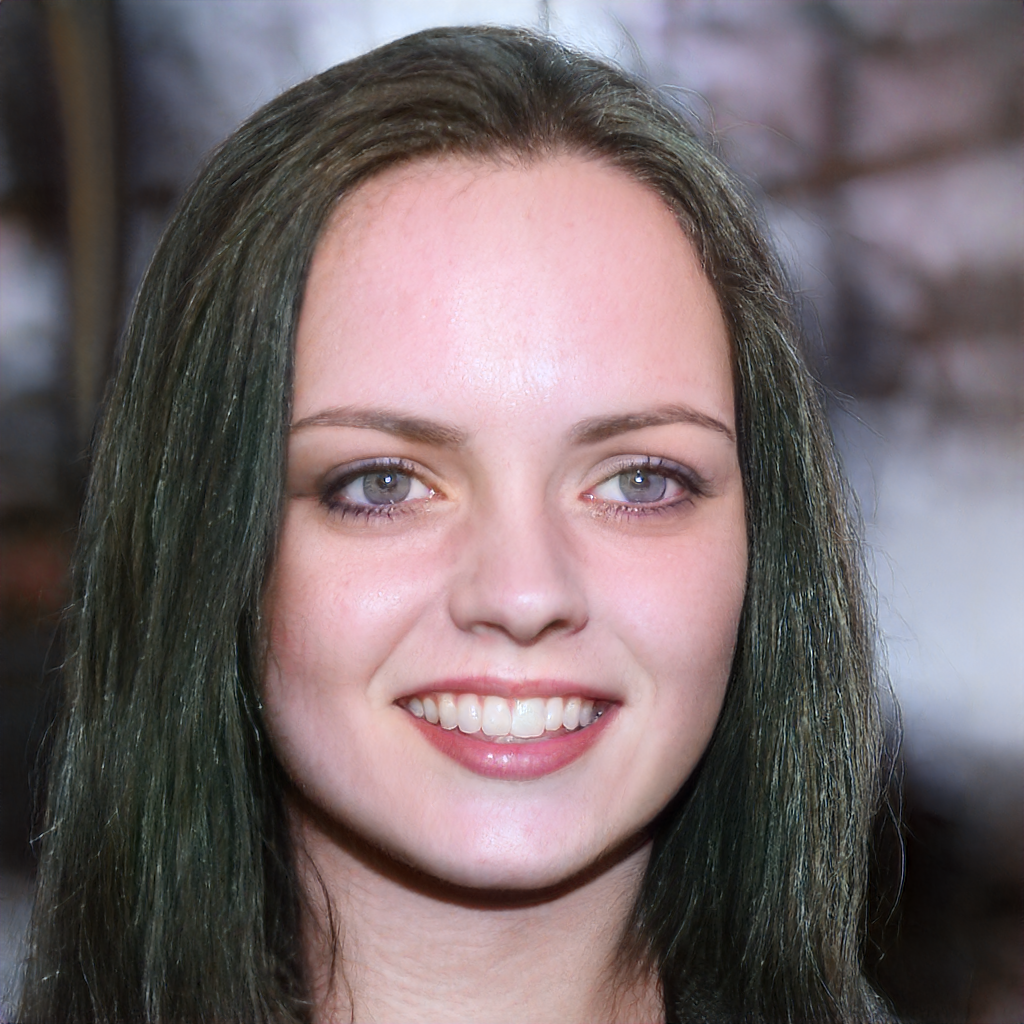}}
    \subfloat{\includegraphics[width=0.203125\columnwidth]{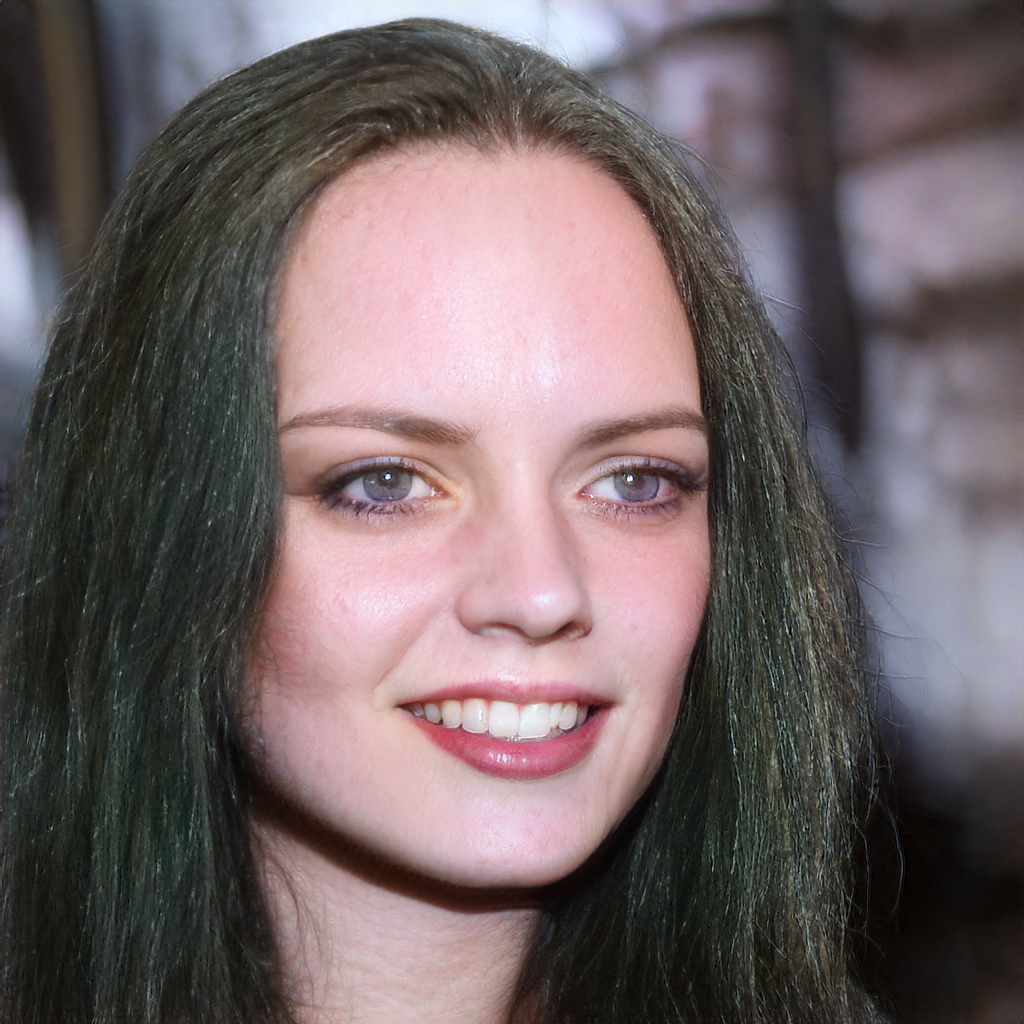}}\\[-2.5ex]
    
    \subfloat{\includegraphics[width=0.203125\columnwidth]{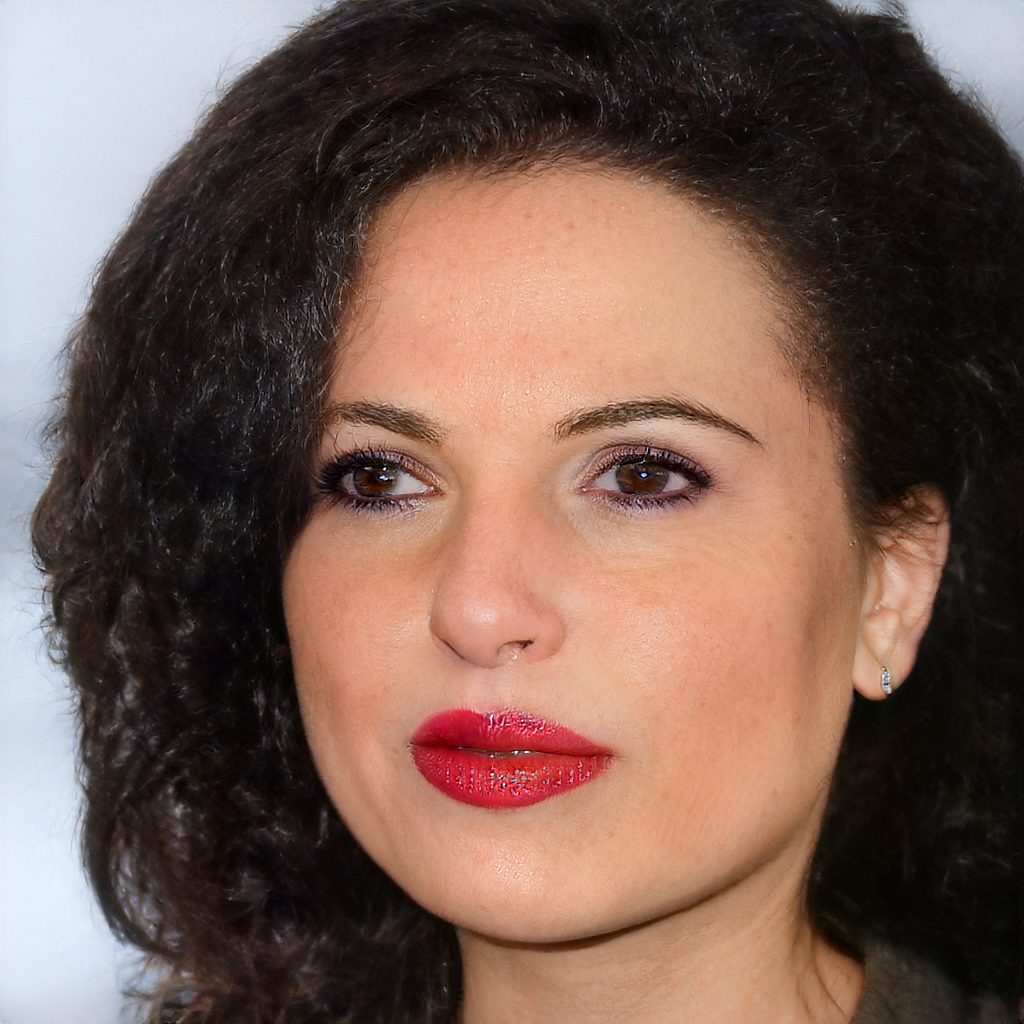}}
    \subfloat{\includegraphics[width=0.203125\columnwidth]{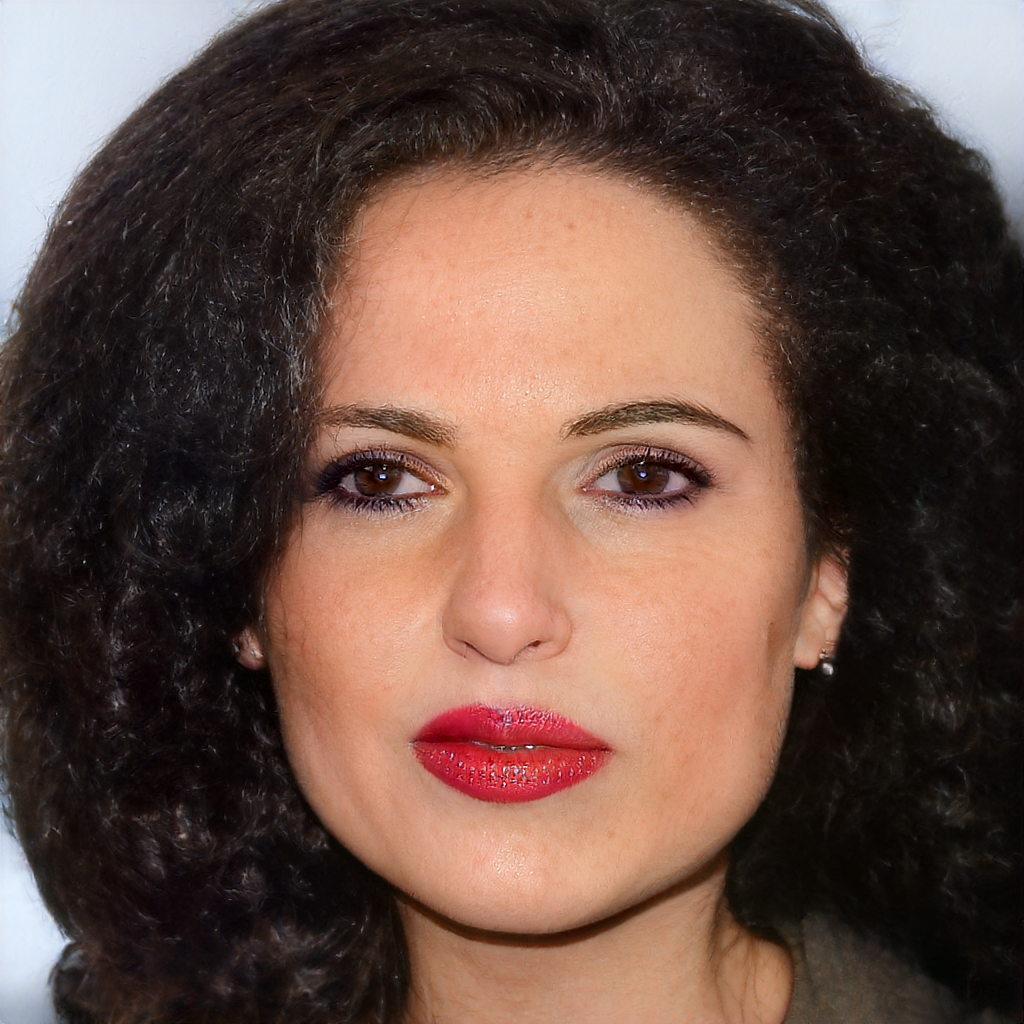}}
    \subfloat{\includegraphics[width=0.203125\columnwidth]{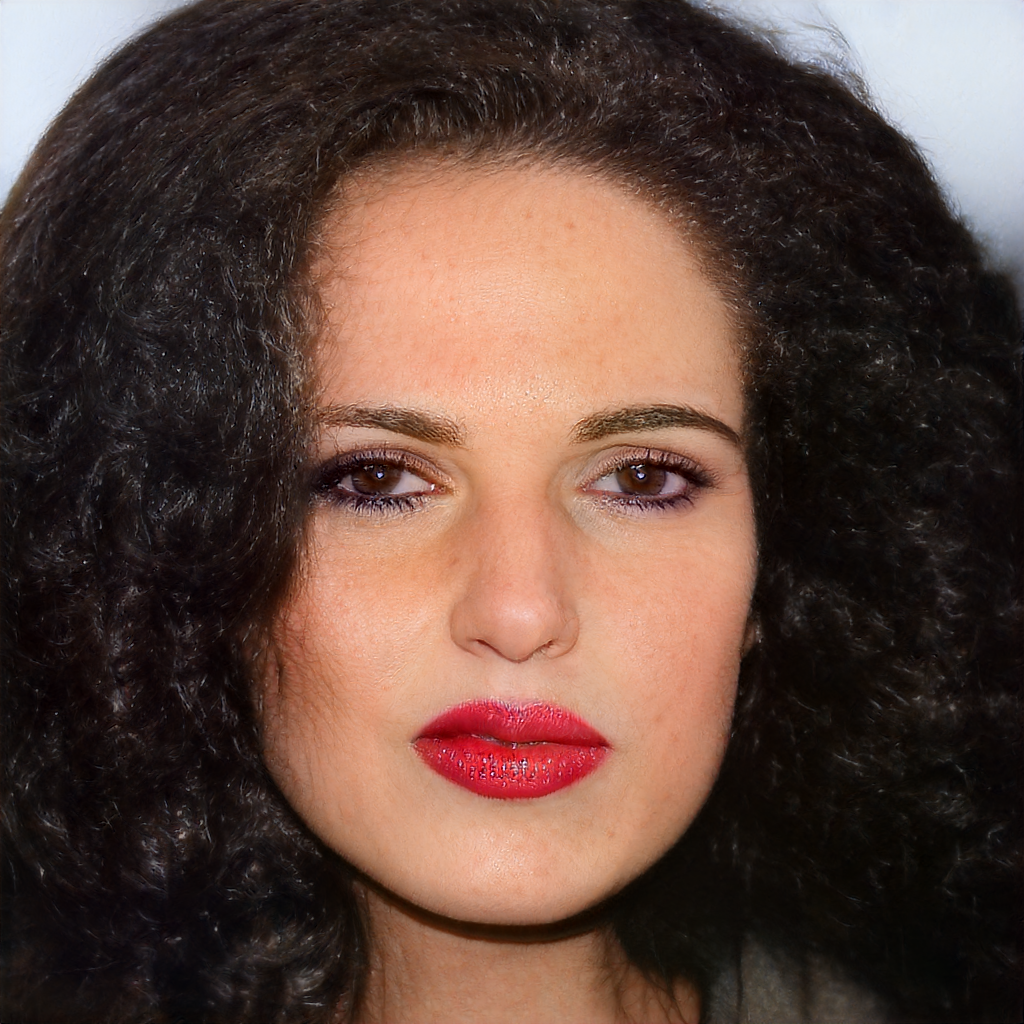}}\\
    
    \caption{Applying precise $\bb{\Delta}$-s to projections of real people images of head pose angles, without any pre-trained GAN supervision.}
    \label{fig:pose control real}
\end{figure}

\begin{figure}
    \centering
    \includegraphics[width=0.65\columnwidth]{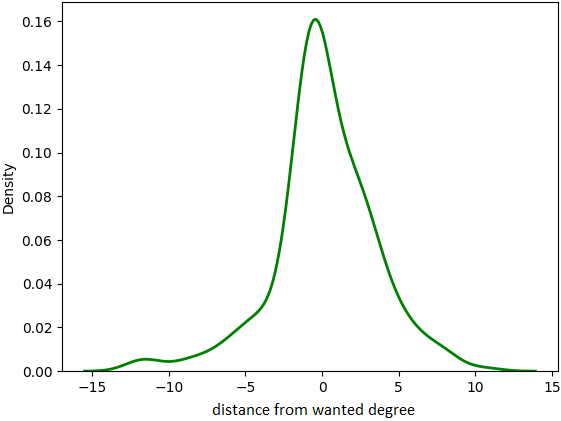}
    \caption{Density graph of angles error across all different angels $\bb{\Delta}$-s in the range of $[-30\degree, 30\degree].$ mean error of $0.15\degree$, and standard deviation of $3.37\degree$.}
    \label{fig:density of errors}
\end{figure}

Another benefit of our Sim2Real model is controlling the exact degree of posture change over randomly synthesized (Fig. \ref{fig:pose control synthesized}) and real-world images (Fig. \ref{fig:pose control real}). Thus, we achieve supervised behavior over an unsupervised GAN in terms of pose control. 
We measure the output face pose degree using \cite{ruiz2018finegrained}, and reach mean degree error of $0.15\degree$ and standard deviation of $3.37\degree$ over changes in range $[-30\degree, 30\degree]$ (Fig. \ref{fig:density of errors}).\\

\subsection{Method Limitations}
\begin{figure}[]
\centering
\subfloat[Additional unwanted changes: Hair growth and eye opening.]{\includegraphics[width = 0.65\columnwidth]{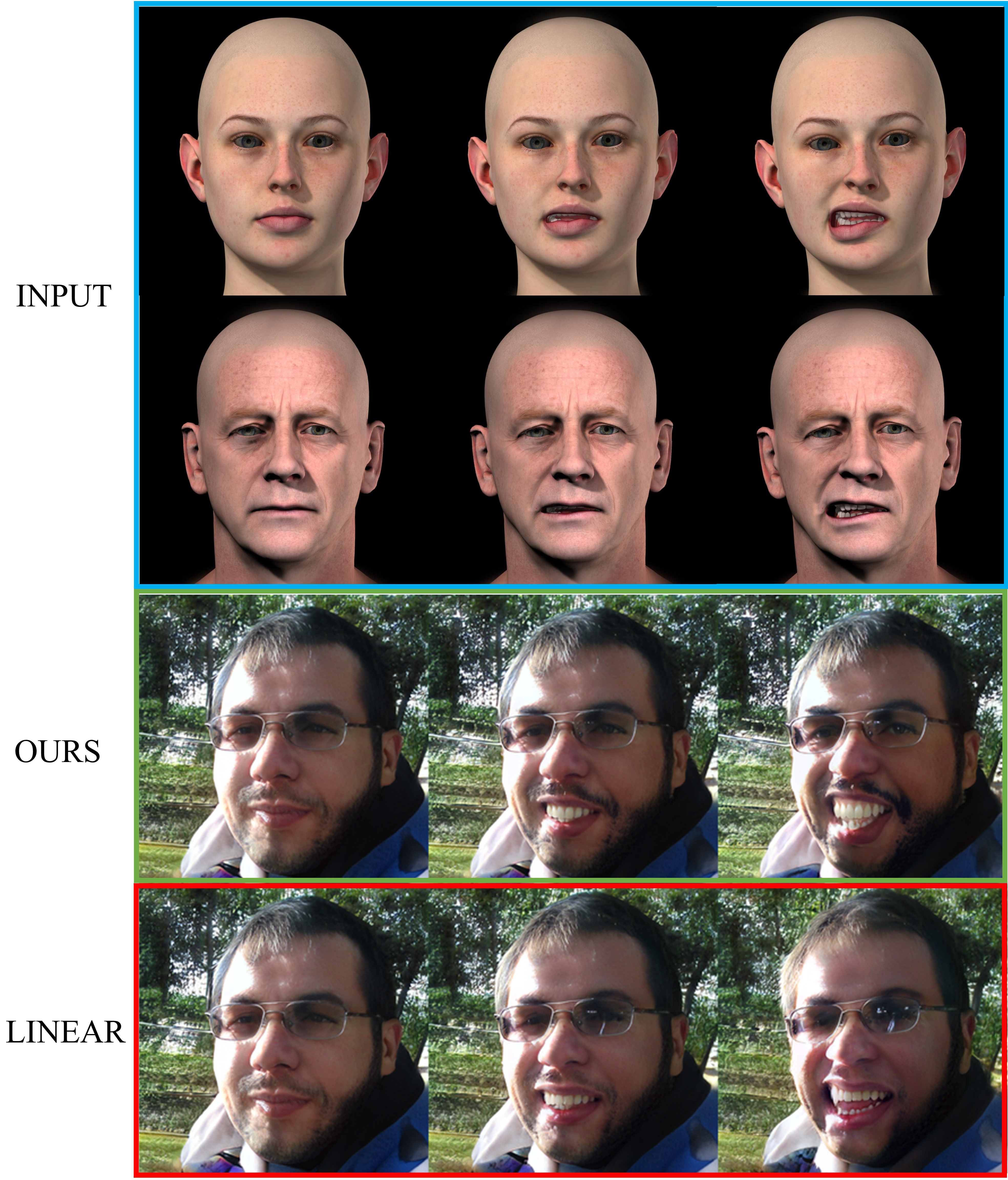}}\\
\subfloat[Additional unwanted changes: Hair style, mouth posture change, eye color and structure change.]{\includegraphics[width = 0.65\columnwidth]{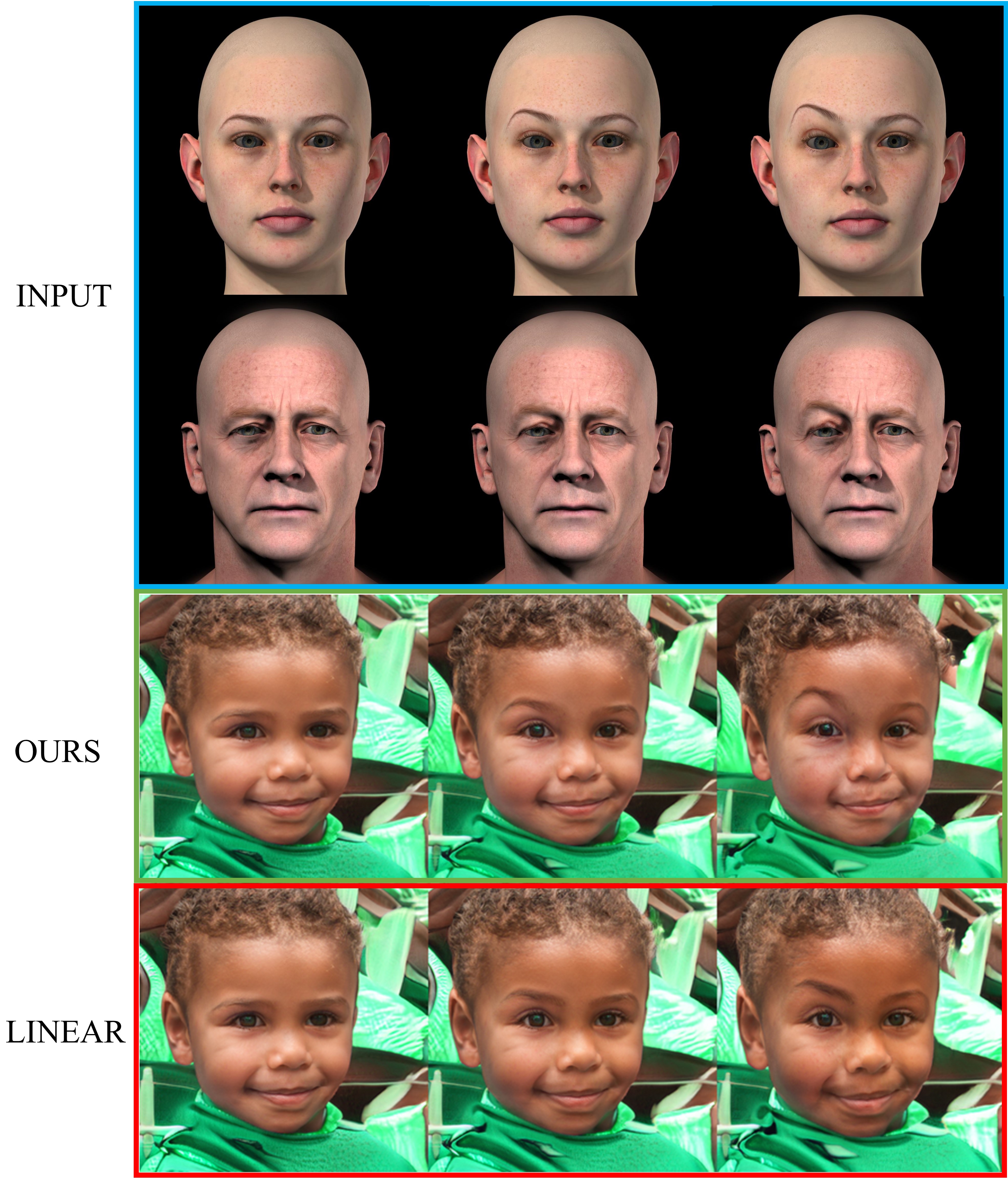}}\\
\subfloat[Additional unwanted changes: Hair color, face hue, face width.]{\includegraphics[width = 0.65\columnwidth]{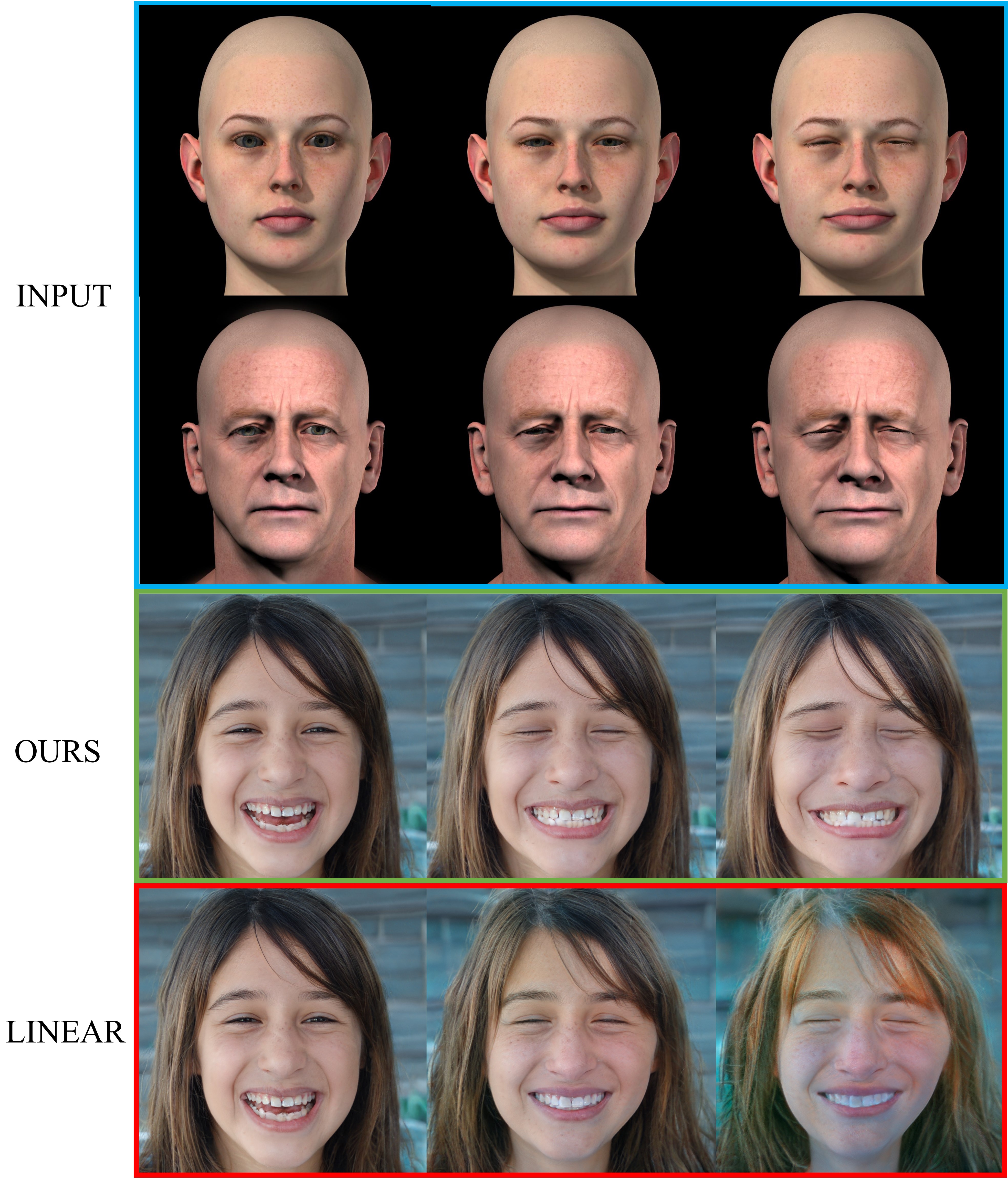}}
\caption{Failure cases of our method.}
\label{figure:Failure cases}
\end{figure}

Lastly, we show several cases where our model fails to achieve its exact goal, resulting in some entangled attribute change. 
There might be several reasons for failing to achieve the goal image:
\begin{enumerate}
    \item Imperfect projection of images into the latent space in terms of semantic meaning.
  \item Uncommonness of expressions in the GAN's training data.
\end{enumerate} 
The failure cases might indicate that some attributes demand more strenuous effort to disentangle than others - as shown in Fig. \ref{figure:Failure cases}.

\begin{table}[]
\captionof{table}{FID score (Frechet Inception Distance) to compare realism of outputs of real images edits between different methods.} \label{tab:fid score} 
\centering

\begin{tabular}{c c c c} 
 \hline
 Editing Method & Light & Pose & Expression \\ [0.8ex] 
 \hline 
 I2S & 31.08 & 31.15 & 56.04 \\ [0.3ex]
 RS-PSP & 24.38 & 27.29 & 48.13 \\[0.3ex]
 Ours & \textbf{17.18} & \textbf{22.48} & \textbf{19.25} \\[0.3ex]

 \hline
 \end{tabular}\\[4ex]

\captionof{table}{Cosine Similarity between the feature vectors of the edited images along a sequence of editing of a specific attribute. A comparison between different methods across different attributes edits is shown in the table below} \label{tab: cosine similarity}

\centering

\begin{tabular}{c c c c} 
 \hline
 Editing Method & Light & Pose & Expression \\ [0.8ex] 
 \hline 
 I2S & 0.91 & 0.877 & 0.941 \\ [0.3ex]
 RS-PSP & 0.939 & 0.931 & 0.966 \\[0.3ex]
 Ours & \textbf{0.971} & \textbf{0.952} & \textbf{0.977} \\[0.3ex]

 \hline
 \end{tabular}\\[4ex]
 \begin{tablenotes}
      \small
      \item notations: I2S - Image2StyleGAN ; RS-PSP - ReStyle based on PSP projection method.
    \end{tablenotes}
\end{table}

\subsection{Implementation Details}
Our Auto-encoder-based model is constructed of an almost-symmetric Encoder and Decoder.
The Encoder's first layer size is twice the size of a $\bb{W^+}$ vector flattened, followed by three hidden layers at the size of a pre-defined $\bb{\Delta} \in R^{64X1}$ and another last output layer at the size of 64. 

The Decoder receives an input sized ($\bb{\Delta} \in R^{64X1} + \bb{W^+} \in R^{512X18}$), followed by the same sized amount of hidden layers and the last layer of a $\bb{W^+} \in R^{512X18}$ vector-sized vector, for outputting the edited residual vector.
All intermediate layers are followed by a Leaky-Relu activation layer with p = 0.25 and a Dropout layer with p = 0.2. 
The model is trained with an ADAM optimizer with a learning rate of 1e-4 and weight decay of 1e-5.
We find that after 10k-20k epochs, which last about 5-10 minutes, the model converges, depending on the specific task, on a single RTX 2080 GPU. 

We measure the FID distance with Pytorch-fid and calculate the Cosine Similarity score over the features of the images extracted with a pre-trained ARCFACE model \cite{deng2019arcface}.

Our synthetic data is facial action coding system (FACS) based \cite{ekman1997face}, and projected to latent vectors with ReStyle-encoder \cite{alaluf2021restyle} based on the PSP method. \cite{richardson2021encoding}.

\section{Conclusions and Future Research}
With only a few samples, which differ by a specific attribute, one can learn the disentangled behavior of a pre-trained entangled generative model.
There is no need for exact real-world samples to reach that goal, which is not necessarily feasible. By using non-realistic data samples, the same goal can be achieved thanks to leveraging the semantics of the encoded latent vectors.
Applying wanted changes over existing data samples can be done with no explicit latent space behavior exploration.

Future research may include testing the limits of the distance between the real world and simulation data, e.g., using a sketched-like simulation, or learning the function that maps the distribution of the synthetic data into the distribution of the real-world data.
The concept may not be exclusive to facial images or even images in general, e.g., illustrated rooms can form a dataset to train a model to change specific attributes like furniture, objects, room structure, and more.
The method proposed throughout this work serves as a sound basis for expansion into the investigation of any black-box model's behavior, given an inverse or approximation of an inverse function.

\clearpage
{\small
\bibliographystyle{ieee_fullname}
\bibliography{egbib}
}

\end{document}